\documentclass[10pt,journal,compsoc]{IEEEtran}
\usepackage{multirow}
\usepackage{ragged2e}
\usepackage{ctable}
\usepackage{bbm}
\usepackage{color}
\usepackage{caption}
\usepackage{subcaption}
\definecolor{myperson}{RGB}{0, 176, 240}
\definecolor{myhorse}{RGB}{102, 255, 51}
\definecolor{myfrrcn_s}{RGB}{112, 48, 160}
\definecolor{myfrcn_s}{RGB}{196, 189, 151}
\definecolor{myfrrcn_m}{RGB}{0, 176, 80}
\definecolor{myfrcn_m}{RGB}{0, 32, 96}
\definecolor{myfrrcn_l}{RGB}{192, 0, 0}
\definecolor{myfrcn_l}{RGB}{0, 176, 240}
\definecolor{myred}{RGB}{255, 0, 0}
\definecolor{mygreen}{RGB}{0, 255, 0}
\definecolor{myblue}{RGB}{0, 0, 255}
\ifCLASSOPTIONcompsoc
  \usepackage[nocompress]{cite}
\else
  \usepackage{cite}
\fi
\usepackage[cmex10]{amsmath}
\usepackage{algorithmic}

\usepackage{array}

\ifCLASSOPTIONcaptionsoff
  \usepackage[nomarkers]{endfloat}
 \let\MYoriglatexcaption\caption
 \renewcommand{\caption}[2][\relax]{\MYoriglatexcaption[#2]{#2}}
\fi
\usepackage{url}

\begin{document}
%
\title{DBF: Dynamic Belief Fusion for Combining Multiple Object Detectors}
%
%
%
%

\author{Hyungtae~Lee,~\IEEEmembership{Member,~IEEE,}
        Heesung~Kwon,~\IEEEmembership{Senior Member,~IEEE}
\IEEEcompsocitemizethanks{\IEEEcompsocthanksitem Hyungtae Lee and Heesung Kwon are with the Intelligent Perception Branch, the Computational \& Information Sciences Directorate (CISD), Army Research Laboratory, Adelphi, MD, 20783 USA (e-mail: \{hyungtae.lee,~heesung.kwon\}.civ@army.mil).\protect\\
E-mail: \{hyungtae.lee,~heesung.kwon\}.civ@army.mil
\IEEEcompsocthanksitem © 2022 IEEE. Personal use of this material is permitted. Permission from IEEE must be obtained for all other uses, in any current or future media, including reprinting/republishing this material for advertising or promotional purposes, creating new collective works, for resale or redistribution to servers or lists, or reuse of any copyrighted component of this work in other works.
}
}

%
%

\markboth{IEEE Transactions on Pattern Analysis and Machine Intelligence}
{Lee \MakeLowercase{\textit{et al.}}}

\IEEEtitleabstractindextext{
\begin{abstract}
\justify

In this paper, we propose a novel and highly practical score-level fusion approach called \emph{dynamic belief fusion} ($DBF$) that directly integrates inference scores of individual detections from multiple object detection methods. To effectively integrate the individual outputs of multiple detectors, the level of ambiguity in each detection score is estimated using a confidence model built on a precision-recall relationship of the corresponding detector. For each detector output, DBF then calculates the probabilities of three hypotheses (\emph{target}, \emph{non-target}, and \emph{intermediate state} (\emph{target} or \emph{non-target})) based on the confidence level of the detection score conditioned on the prior confidence model of individual detectors, which is referred to as \emph{basic probability assignment}. The probability distributions over three hypotheses of all the detectors are optimally fused via the \emph{Dempster's combination rule}. Experiments on the ARL, PASCAL VOC 07, and 12 datasets show that the detection accuracy of the DBF is significantly higher than any of the baseline fusion approaches as well as individual detectors used for the fusion.

\end{abstract}

\begin{IEEEkeywords}
Score-level fusion, Late fusion, Object detection, DBF, Dempster-Shafer theory
\end{IEEEkeywords}}

\maketitle

\IEEEdisplaynontitleabstractindextext

%
\IEEEpeerreviewmaketitle

\IEEEraisesectionheading{\section{Introduction}\label{sec:introduction}}

%
%
%
%

\IEEEPARstart{M}{any} current state-of-the-art methods for fusing multiple object detectors are often limited to feature-level fusion that integrates common intermediate features of individual detectors to extract complementary information for further enhancing detection accuracy~\cite{BFernandoCVPR12,XLanCVPR14,PNatarajanCVPR12,HWangCVPR13}. However, as the field of object detection has advanced rapidly~\cite{NDalalCVPR05,PFelzenszwalbPAMI10,TMalisiewiczICCV11,RGirshickPAMI16,RGirshickICCV15,SRenPAMI17,HLeeArxiv17}, many feature-specific detection algorithms, and related fusion algorithms, are quickly becoming obsolete. Hence, there has been an increasing need for fusion methods that can combine object detection algorithms regardless of their feature types and internal structures for data processing.  One effective solution in this case is late fusion, a process that conditions the \emph{confidence} in individual detector outputs on their prior performance, and then intelligently combines the confidence-weighted probabilistic outputs. 

\begin{figure*}[t]
    \centering
    \includegraphics[trim=5mm 5mm 5mm 5mm,clip,width=0.95\linewidth]{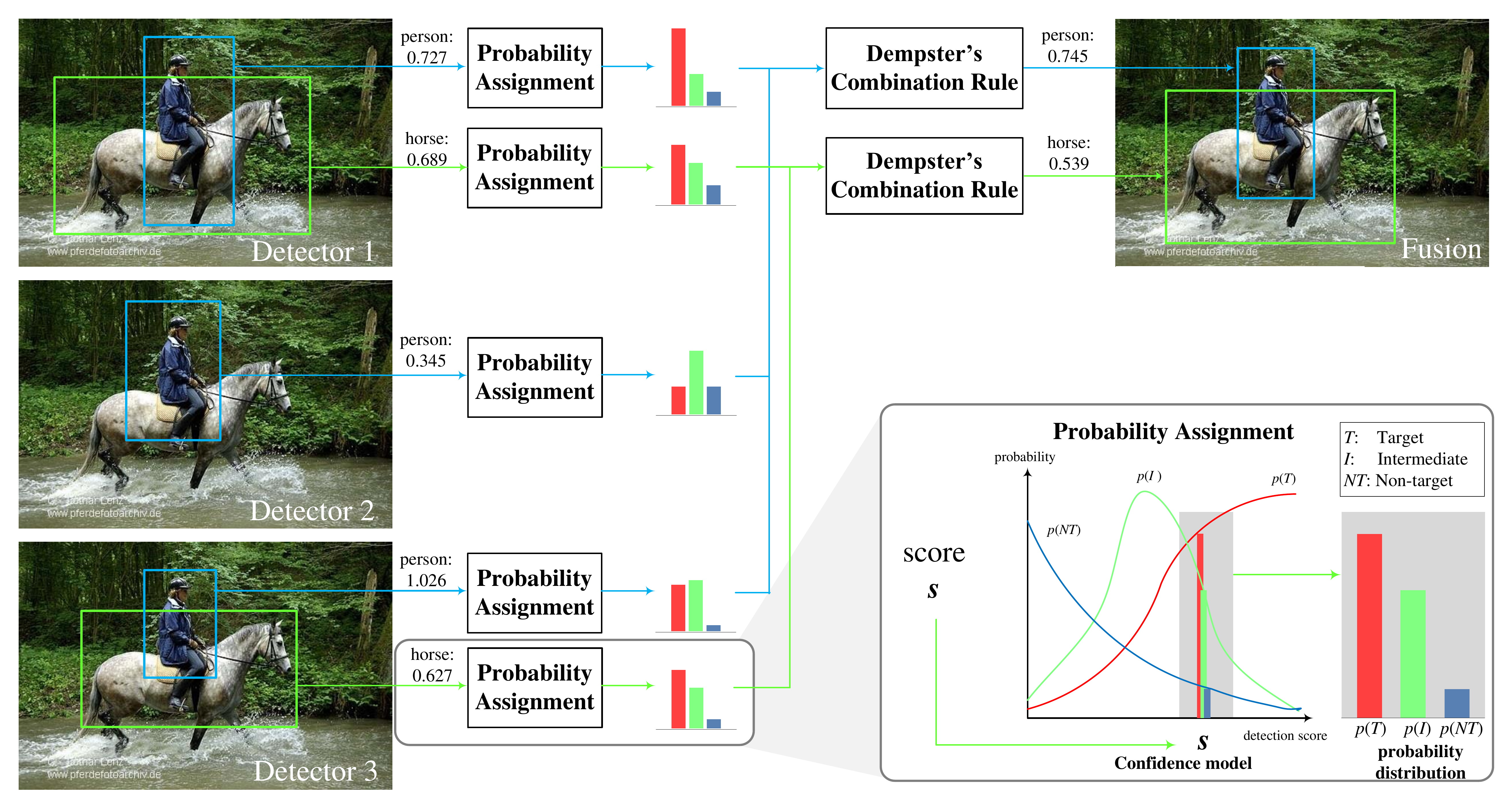}
    \caption{{\bf Dynamic Belief Fusion.} Three detectors detect a \textcolor{myperson}{person} and a \textcolor{myhorse}{horse} in a given image, as shown on the left. For each detection, a score is converted into a probability distribution over three hypotheses (\emph{target}, \emph{non-target}, and \emph{intermediate}) via ``probability assignment''. The confidence model is constructed based on detector's prior performance and formed as a function of a detection score. ``Dempster's combination rule'' combines the probability distributions of all the detectors. This fusion process (the probability assignment and the combination rule) is called Dynamic Belief Fusion (DBF).}
    \label{fig:fusion}
\end{figure*}

In general, the late fusion includes two well-known probabilistic fusion approaches, such as Bayesian fusion and Dempster-Shafer Theory (DST) based fusion, as well as other unstructured or ad hoc approaches. Bayesian fusion is built on Bayesian statistics wherein, in principle, the posterior probabilities of individual hypotheses or events are expressed as the product of joint likelihood estimates of individual detectors and prior beliefs on the hypotheses themselves. However, one major problem of the Bayesian approach is that it cannot inherently leverage and integrate uncertainty or ambiguity associated with individual events into its fusion framework. For example, in Bayesian fusion, there is no way to estimate a degree of belief for ambiguous or uncertain observations (e.g. \emph{target-like} in contrast to \emph{definite-target}), which still includes some evidence of a specific event (e.g. a \emph{target} event) in the observation. In contrast, the belief theory based on DST developed by G. Shafer~\cite{GShaferPrinceton76} takes a step to address ambiguity in observations by considering a compound combination of the hypotheses. When considering two distinctive hypotheses (target and non-target) for object detection, Shafer's belief theory assigns probabilities to the two hypotheses as well as an intermediate state (\emph{target} OR \emph{non-target}) by quantifying the level of ambiguity of the observation that makes either hypothesis more probable. In this manner, a detector output with a high level of ambiguity can be ignored/down-weighted in favor of a more trustworthy, low-ambiguity detector output.  However, assigning these probabilities to all the hypotheses is not a trivial task, and an adequate design of an assignment method is critical to fusion performance.

We propose a novel late fusion approach called Dynamic Belief Fusion (DBF) that dynamically assigns probabilities to all the hypotheses including one associated with uncertainty under the framework of DST. In this approach, confidence in an individual information source (i.e. an individual detector) is estimated by leveraging continuous functions derived from generic detection metrics, such as a precision-recall curve, and is assigned in the form of probabilities to a set of predefined hypotheses for each detection. Multiple object detection algorithms are used as individual information sources (called individual detectors) in DBF for a task of object detection. Figure~\ref{fig:fusion} illustrates the DBF process, in which three heterogeneous detectors generate scores for detection candidates.  Similar to other late fusion methods, these scores are cross-referenced with the detectors' confidence models to obtain a set of probability assignments, essentially re-weighting the outputs of each detector.  Then, the probabilities from multiple detectors are combined into a single fused score via Dempster's combination rule~\cite{ADempsterAMS67}.

To enable continuous probability assignments into \emph{target}, \emph{non-target}, and \emph{intermediate state} hypotheses in the context of object detection, we leverage the precision-recall (PR) model of each detector for building confidence models, from which the probabilities of individual hypotheses are derived. Specifically, to compute the probability associated with the intermediate state, we devise the notion of a theoretical detector, which is a detector that can hypothetically generate detection performance close to a theoretical limit. Unlike a perfect detector, the theoretical detector is subject to performance loss by general detection errors. We assume that, regardless of inference or reasoning approaches for object detection, certain types of errors can always occur even for the theoretical detector, such as a very large-scale training dataset containing images with limited variations in object appearances or different yet very similar looking object categories. To achieve a practical solution, we use an empirical approach in quantifying ambiguity by estimating the PR curves of the theoretical detector and an individual detector. The difference in precision between any individual detector and the theoretical detector at any given observation is then considered as the ambiguity or ignorance in the decision of the individual detector in relation to the theoretical detector. This is because the quantity represents a reduction in accuracy due mainly to deficiencies of the subject detector in terms of algorithms, training data, etc., in comparison with the theoretical detector. Thus, the difference is regarded as the level of ambiguity (or more precisely, ignorance) of the individual detector and the corresponding probability is assigned to the individual detector's intermediate state.

The proposed DBF method is validated on mid- to large-scale datasets, such as ARL~\cite{JTouryanNeuroprosthetics14}, PASCAL VOC 07 and 12~\cite{pascal-voc-2007}. Our preliminary work~\cite{HLeeWACV16} incorporates various plug-in detectors with different levels of detection accuracy on ARL and PASCAL VOC 07. In this paper, to achieve the state-of-the-art detection accuracy, we have carried out additional evaluations with convolutional neural network (CNN)-based object detectors (Fast/Faster R-CNNs~\cite{RGirshickICCV15,SRenPAMI17}) on PASCAL VOC 07 and 12. We also compare DBF to other well-known fusion methods. In these experiments, DBF outperforms all individual object detection methods as well as other fusion methods.

\subsection{Contributions}

Our contributions are summarized as follows:
\begin{enumerate}
	\item We introduce a novel, highly practical, and easy-to-build late fusion framework called Dynamic Belief Fusion that optimally models the joint relationship between the prior information and the current observations of individual plug-in detectors.
	\item DBF dynamically calculates the probabilities for all constituent hypotheses including an intermediate state (\emph{target} or \emph{non-target}) by optimally linking the current detection scores to the prior confidence model of the subject detector derived from the precision-recall relationship estimated based on a validation set.
	\item DBF consistently provides superior performance over the best individual detector as well as all the baseline fusion methods.
\end{enumerate}

\subsection{Differences from Our Preliminary Work~\cite{HLeeWACV16}}

In this paper, we have extensively revised the preliminary work~\cite{HLeeWACV16} by adding:
\begin{enumerate}
    \item Extensive discussion on the background and motivation of the proposed method.
    \item Implementation details of the proposed method so that our methods can be easily rebuilt.
    \item Additional evaluations with current state-of-the-art CNN-based object detectors (Fast/Faster R-CNNs) based on widely used neural network architectures (e.g., VGG, ResNet, etc.) on PASCAL VOC 07 and 12.
    \item Extensive ablation studies to verify the effectiveness of the proposed method and to rationalize the estimation of the theoretical detector's performance in relation to individual detectors.
\end{enumerate}

\section{Related Works}

Fusion of multiple heterogeneous information sources in the literature can be largely divided into two categories: (i) integrating attributes or features generated from various types of methods (early fusion) and (ii) integrating the predicted outputs of multiple methods (late fusion).\medskip

\noindent{\bf Early Fusion.} Kwon and Lee proposed two approaches integrating multiple sample-based tracking approaches using an interactive Markov Chain Monte Carlo (iMCMC) framework~\cite{JKwonCVPR10} and using sampling in a tracker space modeled by Markov Chain Monte Carlo (MCMC) method~\cite{JKwonICCV11}.  Wu et al.~\cite{SWuCVPR14} proposed to combine detectors of different modalities (concept, text, speech, etc) by using relationships among modes in the event detection. Jain et al.~\cite{SJainCVPR17} integrates two stream fully convolutional neural networks, which share similar architecture but take an input of RGB image and the associated optical flow image, respectively. Two streams are integrated into both intermediate and last layers. Eum et al.~\cite{SEumICIP17}, Lee et al.~\cite{HLeeICASSP18}, Dai et al.~\cite{JDaiCVPR16}, Lee et al.~\cite{HLeeICASSP19}, and Lee et al.~\cite{HLeeArXiv19} integrate different machine learning tasks such as object detection, event recognition, and semantic segmentation in a unified convolutional neural network architecture. However, in general, it is not feasible to model the mutual dependencies by fusing multiple approaches built on different principles of extracting or processing attributes or features that represent input data.\medskip

\noindent{\bf Late Fusion (Probabilistic Fusion).} The most popular fusion method based on probabilistic fusion is a Bayesian fusion method.  Manduchi~\cite{RManduchiICCV99} combines textual and color information for image segmentation by Bayesian fusion. Spinello and Siegwart~\cite{LSpinelloICRA08} also use Bayesian fusion as a reasoning rule for fusion of laser range data and camera images for a human detection. Wei et al.~\cite{QWeiJSTSP15} combines remotely sensed multi-band images for scene analysis. Sander and Beyerer~\cite{JSanderSDF13} introduced a variety of applications of Bayesian fusion, followed by its theoretical analysis. However, as previously mentioned, the Bayesian fusion approach cannot inherently leverage the level of uncertainty induced by indistinctive or unclear observation mainly triggered by various deficiencies of the subject detector, which eventually leads to performance degradation.\medskip

\noindent{\bf Late Fusion (Weighted-sum-based Fusion).} Another widely used fusion method is weighted sum (WS). Kim et al.~\cite{THKimICCV13} and Liu et al.~\cite{DLiuCVPR13} used WS methods to fuse multiple types of data for object detection.  For action recognition, Simonyan and Zisserman~\cite{KSimonyanNIPS14} also used the WS method combining predicted scores from two convolutional neural networks that capture the complementary information on appearances from both still frames and motion between frames, respectively. Mees at al.~\cite{OMeesIROS16} uses the WS method that, for object detection, learns optimal weighting of the predictions of different sensor modalities in an online manner. Since the weights are usually optimized by maximizing distance between positive and negative samples, WS also does not provide a way to explain samples from ambiguous observations that do not clearly belong to a positive or a negative class. This triggers fusion performance degradation, as previously mentioned.

To improve upon the late fusion performance, we introduce DBF, a novel fusion framework for object detection that efficiently and dynamically estimates confidence models of individual detectors, from which probabilities for individual hypotheses of each detection are obtained and combined under a DST framework. The main strength of DBF is that it effectively interprets current detection scores by optimally linking them to the associated confidence model and then derives levels of ambiguity in such as way that reinforces evidence of target existence or non-existence of the corresponding observations.  In Section~\ref{sec:exp}, experiments demonstrate superior performance of our proposed approach over WS and Bayesian fusion, as well as other existing methods.

\begin{figure*}[t]
    \centering
    \includegraphics[trim = 5mm 5mm 5mm 5mm,clip,width=0.95\textwidth]{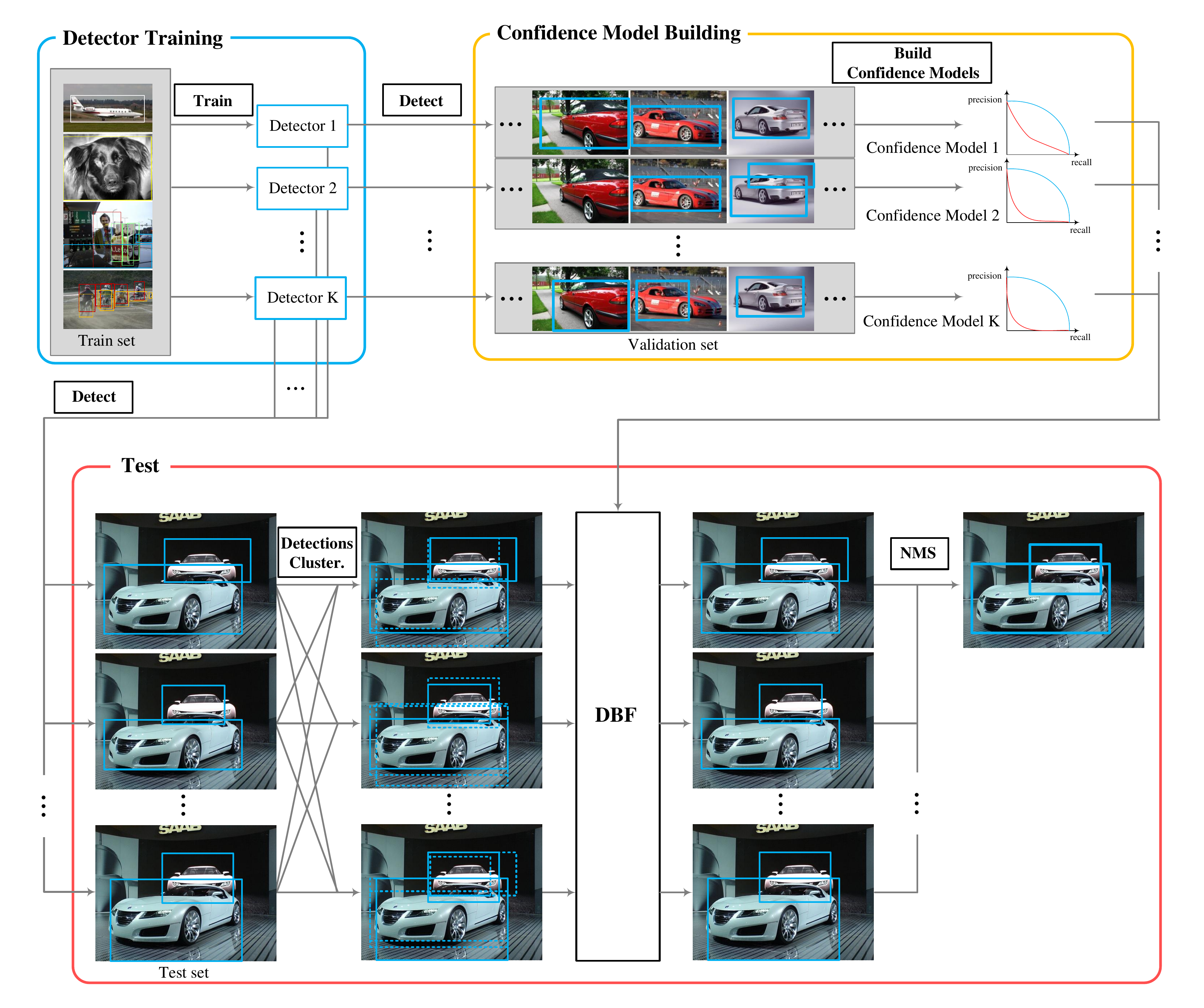}
    \caption{{\bf Flow diagram of the proposed fusion algorithm, DBF.} This diagram shows the fusion process of the `car' category.}
    \label{fig:flow_chart}
\end{figure*}

\section{Background: Dempster-Shafer Theory}
\label{sec:DST}

In this section, we describe the basic components of Dempster-Shafer theory (DST), which form the foundation of our proposed DBF method. Dempster-Shafer theory~\cite{ADempsterAMS67,GShaferPrinceton76} is based on Shafer's belief theory~\cite{GShaferPrinceton76} that obtains a degree of belief for a hypothesis by combining evidences from probabilities of related hypotheses.  DST combines such beliefs from multiple independent sources using a method developed by A. Dempster~\cite{ADempsterAMS67}.

\subsection{Shafer's Belief Theory}

Let $X$ be a universal set consisting of $M$ exhaustive and mutually exclusive hypotheses, i.e. $X = \{H_1, H_2, \cdots , H_M\}$. The power set $2^X$ is the set of all subsets of $X$. (i.e. $2^X=\{\emptyset, H_1, H_2, \cdots, H_M, \{H_1,H_2\}, \cdots, \{H_1,H_2,\cdots,H_M\}\}$) Basic probability in the range $[0,1]$ is assigned to each element of the power set $2^X$. A function defined as $m:2^X \rightarrow [0,1]$ is called a basic probability assignment (BPA). Subsets consisting of compound hypotheses in $X$ represent ambiguity among the constituent hypotheses; the BPA given to the subset measures the level of ambiguity.  A BPA has two properties; (i) $m(\emptyset) = 0$ (the mass of the empty set is zero) and (ii) $\sum_{A\in 2^X}{m(A)=1}$ (the BPA values of the members of the power set sum to one).

From the BPAs, the belief function $bel(A)$ for a set $A$ can be defined as the sum of all basic probabilities which are subsets of the set of interest:
\begin{equation}
bel(A)=\sum_{B|B\subseteq A}{m(B)}.\label{eq:belief}
\end{equation}
$bel(A)$ represents the information in direct support of $A$.

\subsection{Dempster's Combination Rule}

Dempster's combination rule can be applied to calculate a joint BPA from separate BPAs. Under the condition that the evidence from each pair is independent of the other, Dempster's combination rule defines a joint BPA $m_f = m_1 \oplus m_2$, which represents the combined effect of $m_1$ and $m_2$, i.e.,
\begin{equation}
m_f(A)=m_1 \oplus m_2 (A)=\frac{1}{N}\sum_{X\cap Y=A,~A\neq\emptyset}{m_1 (X)m_2 (Y)},
\label{eq:comb_rule_for_two}
\end{equation}
where $N=\sum_{X\cap Y\neq\emptyset}{m_1 (X) m_2 (Y)}$ and $X$ and $Y$ are subsets of $2^X$. $N$ is a measure of the amount of any mass whose common evidence is not the null set. Dempster's rule can be extended to multiple pieces of evidence (e.g., multiple detectors) using the associative and commutative properties of BPAs (i.e. $m_f = m_1\oplus m_2 \oplus\cdots\oplus m_K.$) with the following formula:
\begin{equation}
m_f(A)=\frac{1}{N}\sum_{X_1\cap X_2\cap \cdots \cap  X_K=A}{\prod_{i=1}^K{m_i (X_i)}},
\label{eq:comb_rule}
\end{equation}
where $N=\sum_{X_1 \cap \cdots \cap X_K\neq\emptyset}{\prod_{i=1}^{K}{m_i (X_i)}}$.

\section{The Proposed Fusion Approach}

\subsection{Overview of the Fusion of Detectors}

The proposed fusion process has three phases: (i) individual detector training, (ii) confidence model building, and (iii) the fusion process. These three phases use training/validation/test sets, respectively. Figure~\ref{fig:flow_chart} illustrates the overall process of the proposed DBF. Details of the proposed fusion are as follows.\medskip

\noindent{\bf Detector Training Phase.} $K$ individual detectors are trained on the training set. In this phase, we assume that all the individual detectors are trained to detect a shared set of objects.\medskip

\noindent{\bf Confidence Model Building Phase.} For each detector, a prior confidence model is built by measuring the relationship between target predictivity and sensitivity, referred to as a precision-recall (PR) relationship, based on a validation set to weigh the predicted scores against those of other detectors in terms of individual hypotheses.

In order to calculate the PR relationship, detectors are applied to the images in the validation set to search for potential objects of interest. Each detection candidate forms a pair of a bounding box and an associated detection score. Any detection is considered as {\it positive} if its associated detection score is larger than a certain threshold. All {\it positive} detections are labeled as {\it true} or {\it false positive} by comparing their bounding boxes with groundtruth bounding boxes. Any detection that has an intersection-over-union overlap (PASCAL VOC criteria~\cite{pascal-voc-2007}) of greater than or equal to 0.5 with the groundtruth bounding box is assigned {\it true positive}, otherwise, {\it false positive}.

Precision and recall values can be calculated based on the {\it true/false positive} detections as follows
\begin{itemize}
	\item Precision: $p$ = $\dfrac{N_{TP}}{N_{TP}+N_{FP}}$
	\item Recall: $r$ = $\dfrac{N_{TP}}{N_{tobj}}$,
\end{itemize}
where $N_{TP}$, and $N_{FP}$ are the numbers of true and false positive, respectively.  $N_{tobj}$ is the number of objects of interest. The confidence model is generated from the PR relationship formed by varying thresholds against detection scores of objects on a validation set.\medskip

\noindent{\bf Detection Clustering (Test Phase).} In the test phase, detectors are applied on test images as well.  Let  $d_j^i, i = 1, 2, \cdots, K, j = 1, 2, \cdots, W_i$ be the $j^{th}$ detection of the $i^{th}$ detector, associated with detection score $c_j^i$. For each detection from all the detectors, we collect all the detections from the rest of the detectors that significantly overlap the current subject detection window, which is called ``detection clustering'' (see the second column of the test phase in Figure~\ref{fig:flow_chart}). Two detections $d_j^i$ and $d_l^k$, $i\neq k$ are considered significantly overlapping if the intersection-over-union overlap of their bounding boxes is greater than 0.3.  A $K$-dimensional detection vector ${\bf c} = [c_{j_1}^1~c_{j_2}^2~\cdots~c_{j_K}^K]$ is then constructed, consisting of the scores of the current subject detections and those overlapped from other detectors. If multiple detections from the same detector overlap the current subject detection, the one with the maximum detection score among them is used. If no overlaps exist for a particular detector, the corresponding element of the combined detection vector is filled by a value of $-\infty$ to ignore the influence of the detector in fusion. Note that the number of detection clusters is the same as the number of the total detections from all detectors.\medskip

\noindent{\bf Detection Fusion (Test Phase).} The detection vector of each cluster is used as an input to DBF. Details of the DBF, the main contribution of the proposed work, are described in Section~\ref{ssec:DBF}. DBF calculates the fused detection score for each detection cluster. After scoring all detection clusters by applying DBF, non-maximum suppression is applied to remove duplicated detections from different detectors. All detections, whose intersection over union overlap is greater than 0.3 with any other detection with a high fused detection score, are suppressed.  The final output of the fusion process is a consolidated set of detections, each with a fused detection score.

\subsection{Dynamic Belief Fusion}
\label{ssec:DBF}

\noindent{\bf Object Detection Hypotheses.} For each detection cluster, DBF takes its detection vector ${\bf c}$ as an input and calculates a fused detection score. In object detection, two hypotheses which are \emph{target} and \emph{non-target}, can be considered. We define the universal set $X$ as $\{T,\neg T\}$ and thus its power set is expressed as $\{\emptyset, T,\neg T, \{T, \neg T\}\}$, where $T$ is a \emph{target} hypothesis and $\neg T$ is a \emph{non-target} hypothesis. $\{T, \neg T\}$ in the power set represents detection ambiguity, denoted by $I$ (\emph{intermediate state} or \emph{uncertain state}), which indicates that the subject observation is indistinctive or uncertain and could be either \emph{target} or \emph{non-target}.\medskip

\noindent{\bf Dynamic Basic Probability Assignments.} Based on Shafer's belief theory, we assign basic probabilities to all hypotheses by leveraging the confidence model of each detector. Zero probability is assigned to $\emptyset$ hypothesis. As previously mentioned, we use the PR relationship as the base of confidence models of individual detectors representing their prior performance. We assign basic probabilities to the hypotheses for a given observation (i.e. detection output). Since the PR relationship is obtained by varying a threshold against detection scores, the assigned basic probabilities dynamically change as the threshold changes. Hence, we refer to this assignment as ``dynamic basic probability assignments''.

Precision and recall values with respect to the detection score are used in the dynamic basic probability assignments.  Precisely, when one detection is searched for by a detector, its detection score $c$ is used as the threshold for calculating its recall value ($r$) and the corresponding precision ($p$) is assigned as the basic probability of $\emph{target}$ hypothesis. In DBF, we hypothesize that the remaining portion (i.e., $1-p$) includes latent information about \emph{non-target} and \emph{intermediate hypotheses} and needs to be split into two hidden quantities to account for two basic probabilities associated with the two hypotheses. Note that precision is only defined for targets (objects of interest), not for background objects or non-targets. According to our basic probability assignment rule, the precision of background objects (objects other than the object of interest) should be assigned to the basic probability of non-target. However, the recall of background objects (i.e., recall when ``positive'' refers to background objects) cannot be calculated because the number of the background objects is normally not countable as all the objects or entities other than the target object can be background objects in an input image.

\begin{figure}[t]
    \centering
    \includegraphics[trim = 35mm 85mm 35mm 90mm,clip,width=0.8\linewidth]{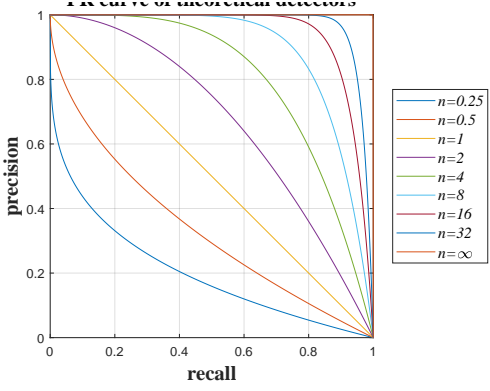}
    \caption{{\bf Precision-recall curve} of detector models with various values of $n$.}
    \label{fig:pr_of_bpd}
\end{figure}

Since the split cannot be achieved based solely on the given PR relationship, we additionally introduce a concept of a theoretical detector, whose performance can possibly achieve a level close to a theoretical limit. We assume that no detector can be perfect or free from error being able to identify all objects of interest without any false alarms (i.e. simultaneously achieving $p=1$ and $r=1$). We define the error portion of the theoretical detector (1 - precision of the theoretical detector) as the basic probability of \emph{non-target} hypothesis. The difference between the precision of an individual detector and that of the theoretical detector is considered as the detection ambiguity (i.e. the probability of the \emph{intermediate state}).

\begin{figure}[t]
    \centering
    \includegraphics[trim = 5mm 5mm 5mm 5mm,width=\linewidth,clip]{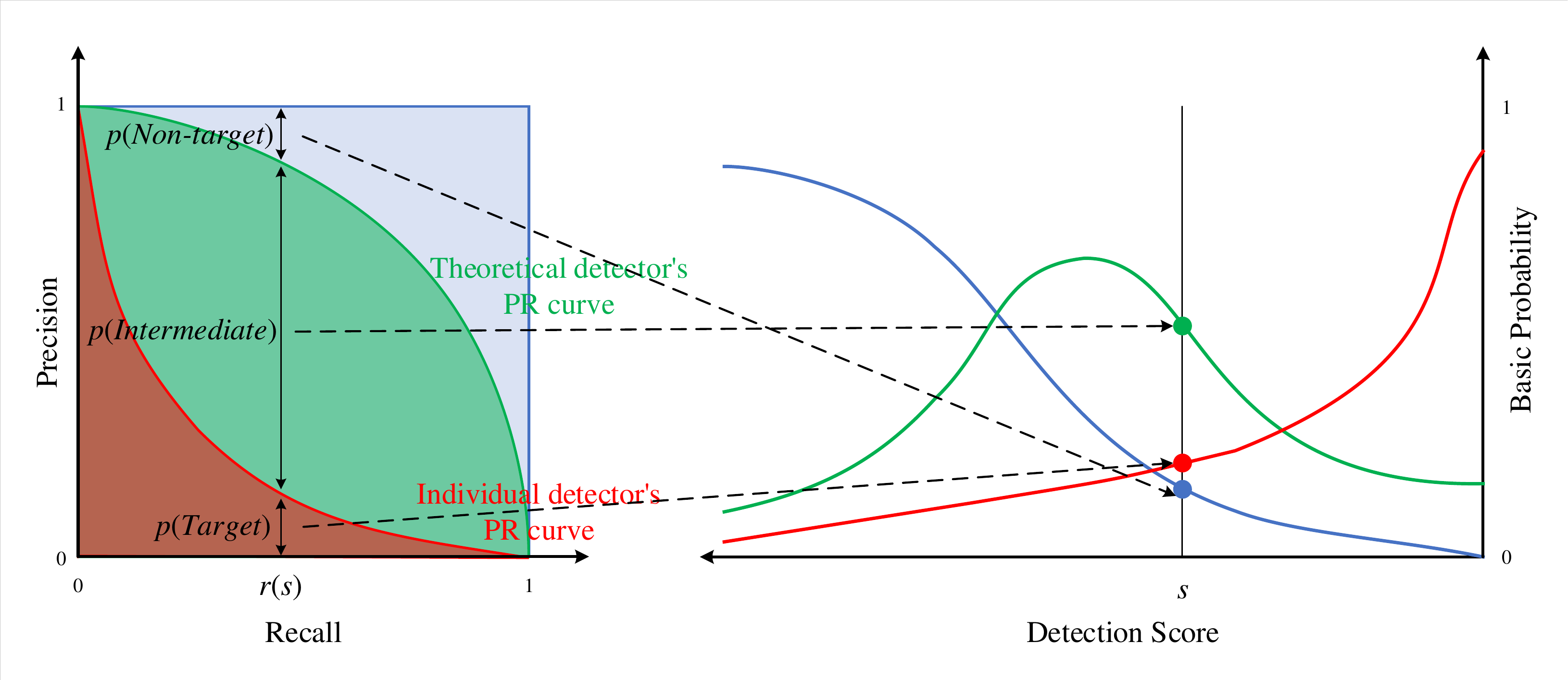}
    \caption{{\bf Dynamic Basic Probability Assignment:} The left plot shows precision-recall curves for both an individual detector and a \emph{theoretical} detector.  The rates of values along the precision axis corresponding to recall $r(s)$ are assigned as the basic probabilities to \emph{target}, \emph{non-target}, and \emph{intermediate state}, where $s$ is a detection score.  The right plot presents the basic probabilities with respect to a detection score, which is derived from the PR curves of the individual detector and the theoretical detector.}
    \label{fig:pr_dba}
\end{figure}

The PR curve of the theoretical detector, $\hat{p}_{bpd}$, is modeled as
\begin{equation}
\hat{p}_{bpd}(r) = 1 - r^n,
\label{eq:fn_bpd}
\end{equation}
where $r$ is recall and $n$ is a positive real number. This model is proposed because in general $\hat{p}_{bpd}$ can mimic the typical behavior of a highly accurate detector, a concave function approaching the top right corner of the plot, such as the car detector in~\cite{PFelzenszwalbPAMI10} when $n$ is larger than one. Figure~\ref{fig:pr_of_bpd} shows the PR curves of different detectors with different parameter $n$. $m(I)$ is defined by $\hat{p}_{bpd} - p$ and the remaining fraction of precision $1-\hat{p}_{bpd}$ is assigned to $m(\neg T)$. Accordingly, given a detection score $c$, the basic probability distribution for three hypotheses is defined as
\begin{eqnarray}
m(T) &\leftarrow& p(c)\nonumber\\
m(\neg T) &\leftarrow& r(c)^n\nonumber\\
m(I) &\leftarrow& 1-p(c)-r(c)^n.\label{eq:prob_assign}
\end{eqnarray}
When $n$ approaches $\infty$, the theoretical detector becomes a perfect detector. Dynamic basic probability assignment is shown in Figure~\ref{fig:pr_dba}. \emph{n} is determined after carrying out cross-validation on validation set. \emph{n} may be different for different object categories.\medskip

\noindent{\bf Fusion.} Fusion of the detections from multiple individual detectors is achieved by computing fused basic probability assignments of $\emph{target}$ and $\emph{non-target}$ hypotheses, $m_f(T)$ and $m_f(\neg T)$, by Dempster's combination rule in Equation~\ref{eq:comb_rule}. The overall fusion score $c_f$ is given by $c_f = bel(T) - bel(\neg T)$ where in our experiments, $bel(T)$ and $bel(\neg T)$ are actually $m_f(T)$ and $m_f(\neg T)$, respectively, according to Equation~\ref{eq:belief} since ${T}$ and $\neg T$ are sets of a single element.

\subsection{Implementation Details}
\label{ssec:implementation_details}

\noindent{\bf Building a Lookup Table for the Confidence Model.} To achieve computational simplicity in the dynamic basic probability assignment in the test phase, the basic probability distribution is pre-computed over the entire range of detection scores in building the confidence models. Let ${\bf c}$ be an array consisting of detection scores of positive detections in a validation set (i.e. ${\bf c} = [c_1,~c_2,\cdots,~c_P]$, where $P$ is the number of positive detections). For each $c$, its precision and recall are calculated by counting {\it true} and {\it false positives} in a validation set. Detection labels are decided according to aforementioned criteria. Then, the basic probability distribution is calculated with the precision and recall values as shown in Equation \ref{eq:prob_assign}. The lookup table consists of a set of detection scores ${\bf c}$ and their corresponding basic probability distributions $\{\{m_1(T),m_1(\neg T),m_1(I)\},~\{m_2(T),m_2(\neg T),m_2(I)\},\cdots,\\\{m_P(T),m_P(\neg T),m_P(I)\}\}$.

Given a detection score $c$, we use linear interpolation to calculate its corresponding basic probability distribution. First, we select two bins, $i$ and $j$ from the lookup table, as follows
\begin{align}
i &= k~~s.t.~~c_k~=~\max(\{ c_l~|~c_l \leq c,~c_l\in{\bf c}\}),~c_k\in {\bf c},\nonumber\\
j &= k~~s.t.~~c_k~=~\min(\{ c_l~|~c_l > c,~c_l\in{\bf c}\}),~c_k\in {\bf c}.   
\end{align}
Then, the corresponding basic probability distribution is calculated as
\begin{equation}
m(H) = \frac{(c_j-c)m_i(H)+(c-c_i)m_j(H)}{c_j-c_i},
\end{equation}
where $H$ can be any hypothesis in $\{T,\neg T,I\}$. Note that each detector has different performance with respect to the precision and recall relationship, so a separate lookup table is required to represent each detector's performance.\medskip

\noindent{\bf Efficient Dempster's Combination Rule.} For each detection cluster ${\bf d} = [d^1,~d^2,\cdots,~d^K]$, its fused probability distribution is calculated over all the cluster elements via the Dempster's combination rule as shown in Equation~\ref{eq:comb_rule}. However, this yields the computation cost, which exponentially increases as the number of detectors, $K$ (i.e. $O(log K)$) increases. It is computationally prohibitive when we use a large number of detectors. 

Note that the Equation~\ref{eq:comb_rule} can be rearranged as:
\begin{eqnarray}
    m_f & = & m_1 \oplus m_2 \oplus m_3 \cdots \oplus m_K\nonumber\\
    & = & (\cdots((m_1 \oplus m_2) \oplus m_3) \cdots \oplus m_K).
\end{eqnarray}
The order of elements can be arbitrarily chosen from a set of clustering detections (i.e. $m_i = m(d^j) \in {\bf d}, i = j$ or $i\neq j$). This can be made possible due to the commutative and associate properties of Dempster's combination rule. 

We can use an efficient method with computation cost $O(K)$ by using the following rearrangement. The method is an iterative process calculating with only two elements at each iteration until all the elements in the detection cluster set are considered. Specifically, two elements are randomly selected from the set ${\bf c}$ and their fused probability distribution $m_f^{(2)}$ are calculated as an initial step. In $t^{th}$ step, the fusion probability distribution $m_f^{(t+1)}$ is recalculated with the pre-computed probability distribution $m_f^{(t)}$ and another element randomly selected from the remaining set. The second step repeats until no element left in the set ${\bf c}$. This efficient combination rule can be expressed as:
\begin{eqnarray}
    m_f^{(2)} & = & m_1 \oplus m_2, \nonumber\\
    m_f^{(3)} & = & m_f^{(2)} \oplus m_3, \nonumber\\
    & \vdots & \nonumber\\
    m_f^{(K)} & = & m_f^{(K-1)} \oplus m_K. 
\end{eqnarray}

For object detection, Dempster's combination rule with two elements, $m_1$ and $m_2$ (equation~\ref{eq:comb_rule_for_two}) can be expresses as
\begin{eqnarray}
    m_f(T) & = & \frac{1}{N}( m_1(T)m_2(T) \nonumber\\
    & & ~~~~~ + m_1(T)m_2(I) + m_1(I)m_2(T) ), \nonumber\\
    m_f(\neg T) & = & \frac{1}{N}(m_1(\neg T)m_2(\neg T) \nonumber\\
    & & ~~~~~ + m_1(\neg T)m_2(I) + m_1(I)m_2(\neg T) ), \nonumber\\
    m_f(I) & = & \frac{1}{N}{m_1(I)m_2(I)},
\end{eqnarray}
where the normalization term $N$ is $m_f(T) + m_f(\neg T) + m_f(I)$.\medskip

\noindent{\bf Bounding Box Refinement.} In conjunction with the detection score fusion process, we also calculate a fused bounding box using bounding box information of all elements in the detection cluster. Denote bounding boxes as $bb_0,~bb_1,\cdots,~bb_K$ associated with the detections of the detection cluster ${\bf d} = [d_1,~d_2,\cdots,~d_K]$, where a bounding box $bb$ consists of coordinates of top-left ($(x_1,~y_1)$) and bottom-right ($(x_2,~y_2)$) corners of the box. Recall that their corresponding detection scores are denoted as ${\bf c} = [c_1,~c_2,\cdots,~c_K]$. A detection score $c_i$ can be converted to the precision value $p_i$ based on the confidence model of its corresponding detector. For each detection cluster, its fused bounding box is calculated with precision values associated with all the detections in the cluster as below:
\begin{equation}
    bb_f = \frac{\sum_{i=1}^{K}{p_i\cdot bb_i\cdot\mathbbm{1}(d_i)}}{\sum_{i=1}^{K}{p_i\cdot\mathbbm{1}(d_i)}},
\end{equation}
where $\mathbbm{1}(d_i)$ is an indicator function that the $i^{th}$ detector contributes its detection to the cluster.

\section{Experiments}
\label{sec:exp}

\subsection{Datasets}

\begin{figure}[t]
    \centering
    \includegraphics[trim = 5mm 10mm 5mm 0mm,clip,width=\linewidth]{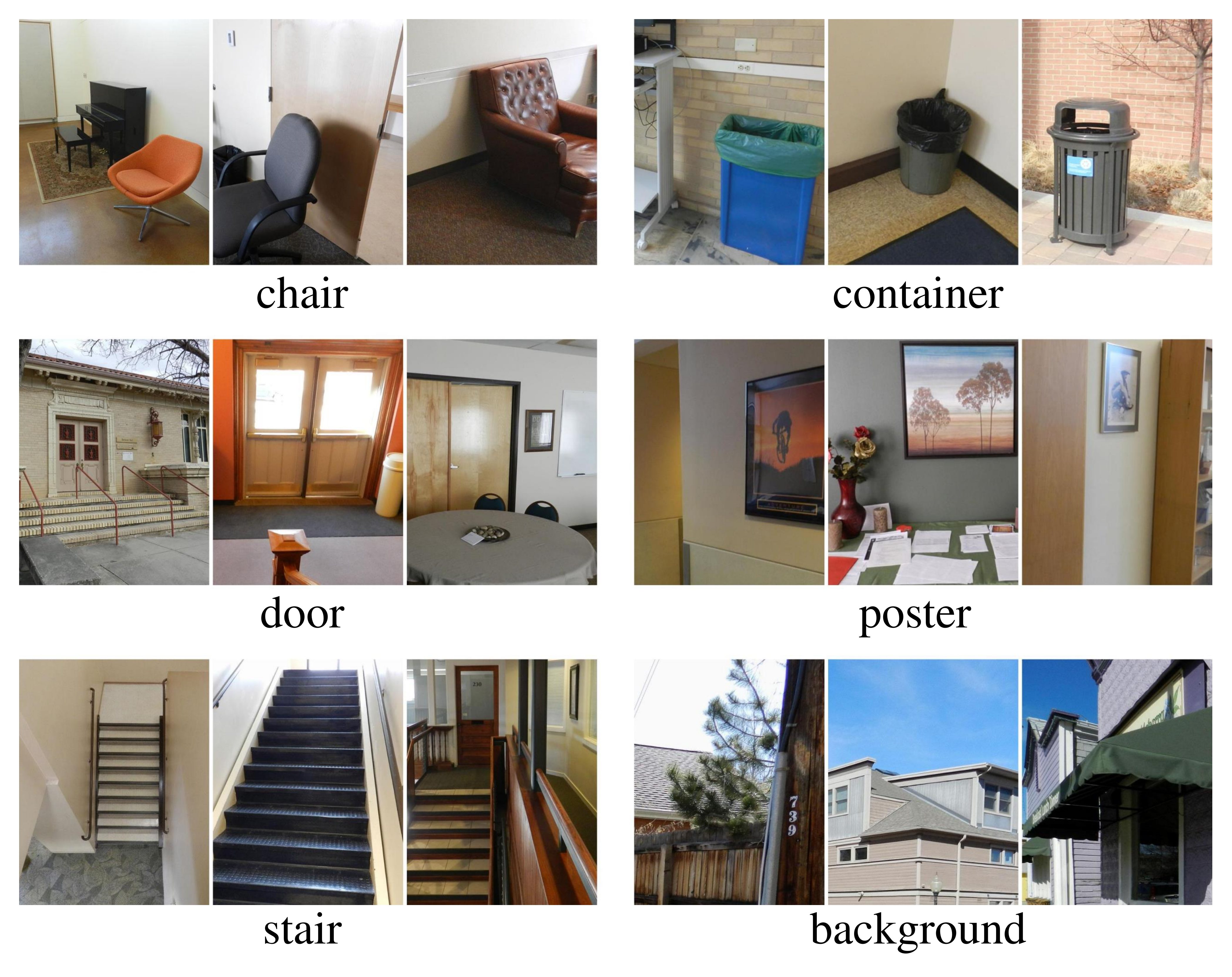}
    \caption{{\bf ARL dataset~\cite{JTouryanNeuroprosthetics14}.}}
    \label{fig:arl}
\end{figure}

\noindent{\bf ARL Dataset.} The Army Research Lab (ARL) image dataset was originally created for the purpose of analyzing human performance in Rapid Serial Visual Presentation (RSVP)~\cite{JTouryanNeuroprosthetics14} tasks, but is also applicable to object detection tasks.  (In \cite{RRobinsonIROS15,HLeeIROS16}, we integrated computer vision-based object detection with human decisions on this dataset.)  The dataset contains 3000 images of both indoor and outdoor scenes, 1438 images of which contain at least one object-of-interest. The target objects include: chair, container, door, poster, and stair.  We randomly select 971/596/1533 images from the ARL dataset and assign them into {\tt train}, {\tt val}, and {\tt test} sets, respectively. Figure~\ref{fig:arl} displays several example images of all five objects as well as background images, the images with no objects of interest.  The number of images in the ARL dataset is relatively small compared to that of other benchmark datasets, such as PASCAL VOC datasets, but, with regard to the mean average precision (mAP), the ARL dataset (0.253 for DPM) is not considerably less challenging compared to the benchmark datasets (0.239 for DPM on PASCAL VOC 07).\medskip

\noindent{\bf PASCAL VOC 07 \& 12 Datasets.} We also used PASCAL VOC 07 and 12, which are widely used object detection benchmark datasets. PASCAL VOC protocol provides {\tt train}, {\tt val}, {\tt trainval}, and {\tt test}, where the {\tt trainval} set consists of images of {\tt train} and {\tt val} sets. For our experiment, we use two different partitions: {\tt train}/{\tt val}/{\tt test} and {\tt trainval}/{\tt trainval}/{\tt test}. The first and second sets of each partition are used for training detectors and building the confidence model, respectively, while the third is used for testing the method and its evaluation. The first partition is made to avoid the confidence model overfitting the training set. Therefore, the performance of the individual detectors used by the first partition is worse than the performance reported in the original literature with regard to the individual detectors, as we are using a smaller training dataset.

\subsection{Evaluation Methods}

We have prepared three different settings to evaluate the effectiveness of the proposed fusion method.

{\bf The first setting} is to integrate detections from multiple weak detectors on the ARL dataset. A weak detector is a method, which generally provides relatively poor performance when compared with CNN-based detectors.

{\bf The second setting} is to adopt a range of different detection algorithms with different detection accuracy, from weak detectors to CNN-based detectors, on PASCAL VOC 2007. This setting used the first dataset partition of {\tt train}/{\tt val}/{\tt test}.

{\bf  The third setting} is used to evaluate multiple advanced CNN-based detection methods on both PASCAL VOC 2007 and 2012. This evaluation used the second dataset partition of {\tt trainval}/{\tt trainval}/{\tt test}.

The first and second evaluation settings are previously used in our preliminary paper~\cite{HLeeWACV16} and the third setting is carried out first in this paper.

\subsection{Evaluation of ARL dataset}

\subsubsection{Individual Detectors}

We have selected four object detectors with different principles of processing input images to detect objects of interest whose codes are readily available online are selected: Dense SIFT (DSIFT)~\cite{CLiuPAMI11}, Transductive Annotation by Graph (TAG)~\cite{JWangICML08}, Examplar SVM (ESVM)~\cite{TMalisiewiczICCV11}, and Deformable Part Models (DPM)~\cite{PFelzenszwalbPAMI10}.\medskip

\noindent{\bf DSIFT.} SIFT~\cite{DLoweIJCV04}, from which DSIFT is originated, is a local image descriptor and has two components: feature extraction and feature description. DSIFT uses a densely sampled grid instead of the region selection based on the SIFT feature extraction and computes the SIFT feature description for each grid. Image representation is built by concatenating the descriptors from all the grids of the image. SVM is used as an object classifier.\medskip

\noindent{\bf TAG.} TAG is a graph-based label propagation method using a small set of labeled images to derive likely labels based on image similarity metrics. TAG-based object detection method is trained in a semi-supervised fashion that, for training, uses randomly selected 5\% images with object category labels and the remaining images without labels. TAG is used to estimate the labels of the remaining images.\medskip

\noindent{\bf ESVM.} ESVM learns a SVM-based separate classifier for each positive training image (called as an exemplar) using a Histogram of Orientied Gradient (HOG) feature~\cite{NDalalCVPR05}, and scores candidate detection hypotheses based on ``distance'' to exemplars.\medskip

\noindent{\bf DPM.} DPM represents objects as sets of parts that can be deformed using HOG features at two scales and latent features, with a deformation cost. The latent SVM is used to optimize this representation and output a score for each object location hypothesis.

\begin{table}[t]
\caption{{\bf Average Precision on the ARL dataset~\cite{JTouryanNeuroprosthetics14}.} For each object category, best accuracies among all individual detectors and among all fusion methods are depicted in a {\bf bold} font. Best accuracy among all methods are depicted by an \underline{underline}.}
\setlength{\tabcolsep}{7.5pt}
\renewcommand{\arraystretch}{1.4}
\centering
\begin{tabular}{l|c|ccccc}
\specialrule{.15em}{.05em}{.05em} 
& \multirow{2}{2.5em}{mAP} & \multicolumn{5}{c}{AP per category} \\
& & chair & container & door & poster & stair \\\specialrule{.15em}{.05em}{.05em} 
DSIFT & .091 & .143 & .037 & .073 & .143 & .061 \\
TAG & .082 & .045 & .128 & .165 & .066 & .008 \\
ESVM & .190 & .125 & .318 & .150 & .236 & .122 \\
DPM & {\bf .253} & {\bf .188} & {\bf .396} & {\bf .194} & {\bf .342} & {\bf .143} \\\specialrule{.15em}{.05em}{.05em} 
Platt & .238 & .191 & .364 & .204 & .307 & .125 \\
WS & .252 & .192 & .388 & .267 & .318 & .096 \\
Bayes & .276 & .244 & .424 & .281 & .341 & .089 \\\hline
DBF & \underline{{\bf .325}} & \underline{{\bf .329}} & \underline{{\bf .451}} & \underline{{\bf .298}} & \underline{{\bf .390}} & \underline{{\bf .159}} \\\specialrule{.15em}{.05em}{.05em} 
\end{tabular}
\label{tab:map_arl}
\end{table}

\begin{table*}[t]
\caption{{\bf Average Precision on the PASCAL VOC 07 dataset.}}
\setlength{\tabcolsep}{3.7pt}
\renewcommand{\arraystretch}{1.4}
\centering
{\scriptsize
\begin{tabular}{l|c|cccccccccccccccccccc}
\specialrule{.15em}{.05em}{.05em}  
& \multirow{2}{2.5em}{mAP} & \multicolumn{20}{c}{AP per category}\\
&& aero & bike & bird & boat & bottle & bus & car & cat & chair & cow & table & dog & horse & mbike & persn & plant & sheep & sofa & train & tv \\\specialrule{.15em}{.05em}{.05em} 
HOG & .021 & .036 & .060 & .001 & .001 & .005 & .005 & .094 & .001 & .001 & .092 & .001 & .002 & .002 & .005 & .001 & .001 & .003 & .001 & .013 & .103 \\
TAG & .026  & .019 & .051 & .009 & .002 & .002 & .028 & .022 & .080 & .002 & .00 & .056 & .032 & .020 & .085 & .051 & .002 & .001 & .010 & .020 & .014 \\
DSIFT & .064 & .081 & .024 & .017 & .004 & .002 & .080 & .118 & .142 & .005 & .097 & .109 & .128 & .040 & .037 & .076 & .002 & .059 & .102 & .122 & .028 \\
ESVM & .182  & .164 & .418 & .041 & .096 & .107 & .341 & .336 & .095 & .100 & .129 & .097 & .013 & .362 & .322 & .170 & .033 & .170 & .102 & .287 & .263 \\
Color Attributes & .246  & .201 & .518 & .026 & .102 & .167 & .344 & .363 & .172 & .158 & .198 & .041 & .358 & .349 & .436 & .376 & .106 & .128 & .273 & .304 & .307 \\
DPM & .239  & .231 & .500 & .036 & .099 & .162 & .388 & .451 & .153 & .120 & .172 & .129 & .106 & .463 & .375 & .346 & .109 & .109 & .144 & .353 & .333 \\
Loc-CNN$\dagger$ & .041 & .010 & .080 & .035 & .031 & .001 & .048 & .030 & .074 & .011 & .040 & .039 & .063 & .099 & .078 & .035 & .022 & .022 & .018 & .046 & .034 \\
R-CNN & {\bf .540}  & {\bf .637} & {\bf .709} & \underline{{\bf .506}} & {\bf .393} & {\bf .300} & {\bf .639} & {\bf .721} & \underline{{\bf .601}} & {\bf .303} & \underline{{\bf .585}} & {\bf .458} & \underline{{\bf .559}} & {\bf .631} & \underline{{\bf.681}} & {\bf .549} & {\bf .291} & {\bf .536} & {\bf .467} & {\bf .575} & {\bf .662} \\\specialrule{.15em}{.05em}{.05em}
Platt & .511 & .596 & .695 & .470 & .383 & .314 & .627 & .708 & .566 & .295 & .542 & .398 & .529 & .595 & .640 & .508 & .278 & .503 & .439 & .537 & .605 \\
WS & .516 & .576 & .692 & .486 & .370 & .326 & .601 & .706 & .526 & .315 & .533 & .450 & .511 & .658 & .628 & .538 & .273 & .502 & .466 & .577 & .594 \\
Bayes & .354 & .460 & .616 & .177 & .098 & .297 & .541 & .644 & .252 & .115 & .413 & .278 & .344 & .359 & .517 & .229 & .215 & .447 & .138 & .461 & .475 \\
LEF & .510 & .606 & .671 & .441 & .366 & .291 & .624 & .721 & .503 & .300 & .571 & .444 & .463 & .621 & .615 & .524 & .276 & .503 & .488 & .528 & .628 \\
D2R & .531 & .609 & .687 & .468 & \underline{{\bf .398}} & .311 & \underline{{\bf .665}} & \underline{{\bf .757}} & .552 & .326 & .587 & .449 & .493 & .660 & .636 & .528 & .289 & .511 & .502 & .550 & .654 \\\hline
DBF & \underline{{\bf .553}}  & \underline{{\bf .650}} & \underline{{\bf .720}} & {\bf .501} & .392 & \underline{{\bf .341}} & .658 & .729 & {\bf .576} & \underline{{\bf .339}} & {\bf .578} & \underline{{\bf .477}} & {\bf .537} & \underline{{\bf .670}} & {\bf .664} & \underline{{\bf .572}} & \underline{{\bf .315}} & \underline{{\bf .537}} & \underline{{\bf .539}} & \underline{{\bf.590}} & \underline{{\bf .672}} \\\specialrule{.15em}{.05em}{.05em} 
\multicolumn{22}{p{17.5cm}}{$\dagger$Unexpected low performance for Loc-CNN is due to coarse-grid scanning window strategy with fixed square aspect ratio.}
\end{tabular}}
\label{tab:map_pascal}
\end{table*}

\subsubsection{Baseline Fusion Methods}

As a baseline, we used three approaches: Platt scaling~\cite{JPlattALMC99}, Weighted Sum (WS)~\cite{OMeesIROS16}, and Bayesian fusion~\cite{JSanderSDF13}.\medskip

\noindent{\bf Platt Scaling.} The Platt scaling learns a logistic regression model ($\alpha$, $\beta$) on the detection scores of true and false positive detections. Given score $c$ and the learned model parameters, $\alpha$ and $\beta$, the calibrated detection score is as follows:
\begin{equation}
    f_{Platt}(x|\alpha, \beta)=\frac{1}{1+e^{-\alpha (x+\beta)}}.
\end{equation}
We applied Platt scaling to all the detectors on validation images. At test time, detections from multiple different detectors can be reconciled by fitting the distribution of detection scores of each detector to that of the Platt-scaled validation set. After scaling, the maximum value of the combined detector vector {\bf c} is used as the final fused score.\medskip

\noindent{\bf WS.} The WS method scores a detection cluster with its associated score vector {\bf c} by multiplying it with a weight vector {\bf w} as below:
\begin{equation}
    f_{WS}({\bf c}) = {\bf w}^T {\bf c}.
\end{equation}
The weight {\bf w} is learned through linear SVM optimization (using {\tt LibLinear} library~\cite{RFanJMLR08}). In WS, detection scores are converted into probabilities by Platt scaling as well. This is because negative infinity scores in the detection score {\bf c} can disable the SVM optimization.\medskip

\begin{table}[t]
\setlength{\tabcolsep}{9.7pt}
\renewcommand{\arraystretch}{1.4}
\begin{center}
\begin{tabular}{c|cccccc}
\specialrule{.15em}{.05em}{.05em} 
 & Platt & WS & Bayes & LEF & D2R & DBF \\\specialrule{.15em}{.05em}{.05em} 
mAP & .268 & .271 & .253 & .283 & .261 & {\bf.341} \\\specialrule{.15em}{.05em}{.05em} 
\end{tabular}
\end{center}
\vspace{-0.4cm}
\caption{{\bf Performance of fusion approaches with all the detectors except R-CNN} on the PASCAL VOC 07 dataset.}
\label{tab:map_pascal2}
\end{table}

\noindent{\bf Bayesian Fusion.} For Bayesian fusion, we use a na{\' i}ve Bayesian model assuming that all the approaches are independent of each other. In other words, the joint likelihood can be decomposed as the product of the likelihoods of each detector, while the posterior probability ($m(H|{\bf c})$) is expressed as the product of the prior probability ($m(H)$) and the joint likelihood ($l_k(c_k|H)$, $k=1,2,\cdots,K$)) as below:
\begin{equation}
    m(H|{\bf c})=m(H)\prod_{k=1}^{K}{l_k(c_k|H)},
\end{equation}
where $H$ can be any hypothesis in $\{T,~\neg T\}$. 

For each detector, a score range from the highest detection score to the lowest detection score in validation set was divided equally into 20 bins. $l_k (c_k | H)$ is defined as a rate of (true or false) detections in the bin containing $c_k$ from all (true or false) detections. By equally treating prior probability of target and non-target (i.e. $m(T) = m(\neg T) = \frac{1}{2}$), we consider the fusion score as below:
\begin{eqnarray}
    m_f({\bf c}) & = & m(T|{\bf c}) - m(\neg T|{\bf c})\nonumber\\
    & = & \frac{1}{2}\prod_{k=1}^{K}{l_k(c_k|T)} - \frac{1}{2}\prod_{k=1}^{K}{l_k(c_k|\neg T)}.
\end{eqnarray}

\subsubsection{Detection Accuracy}

Table~\ref{tab:map_arl} shows that DBF outperformed all the baseline fusion algorithms as well as individual detectors on the ARL dataset by at least mAP of .072 and .049, respectively. DBF also provides the best detection accuracy for all the object categories.

\subsection{Evaluation of PASCAL VOC 07}

\begin{table*}[t]
\caption{{\bf Average Precision on the PASCAL VOC 07 dataset.} For this evaluation, six R-CNNs belong to the detector pool.}
\setlength{\tabcolsep}{2.7pt}
\renewcommand{\arraystretch}{1.4}
\centering
{\scriptsize
\begin{tabular}{ll|c|cccccccccccccccccccc}
\specialrule{.15em}{.05em}{.05em} 
\multicolumn{2}{c|}{Backbone} & \multirow{2}{2.5em}{mAP} & \multicolumn{20}{c}{AP per category} \\
img. classif. & obj. detect & & aero & bike & bird & boat & bottle & bus & car & cat & chair & cow & table & dog & horse & mbike & persn & plant & sheep & sofa & train & tv \\\specialrule{.15em}{.05em}{.05em} 
\multirow{2}{5em}{VGG M} & Fast R-CNN & .606 & .687 & .711 & .597 & .445 & .282 & .671 & .738 & .736 & .366 & .676 & .630 & .702 & .747 & .675 & .623 & .295 & .571 & .651 & .710 & .600 \\
 & Faster R-CNN & .607 & .632 & .716 & .580 & .461 & .337 & .649 & .748 & .727 & .380 & .642 & .578 & .673 & .768 & .710 & .672 & .327 & .607 & .571 & .720 & .635 \\\hline
\multirow{2}{5em}{VGG 16} & Fast R-CNN & .686 & .733 & .789 & .681 & .591 & .407 & .781 & .795 & .814 & .478 & .745 & .675 & .800 & .823 & .753 & .724 & .329 & .683 & .682 & {\bf .776} & .654 \\
 & Faster R-CNN & .693 & .681 & .785 & .688 & .561 & .502 & .803 & .795 & .800 & .508 & .756 & .634 & .813 & .830 & .744 & .762 & .390 & .696 & .649 & .755 & .714 \\\hline
\multirow{2}{5em}{ResNet 101} & Fast R-CNN & .718 & {\bf .778} & {\bf .796} & .747 & {\bf .601} & .456 & .794 & .795 & \underline{{\bf .863}} & .537 & .804 & {\bf .699} & .868 & .827 & .758 & .733 & .372 & .719 & {\bf .736} & .758 & .712 \\
 & Faster R-CNN & {\bf .739} & .771 & .790 & {\bf .768} & .578 & \underline{{\bf .588}} & {\bf .835} & {\bf .826} & .857 & {\bf .575} & \underline{{\bf .827}} & .679 & \underline{{\bf .878}} & {\bf .843} & {\bf .788} & {\bf .782} & {\bf .447} & {\bf .737} & .734 & .764 & {\bf .715} \\\specialrule{.15em}{.05em}{.05em} \specialrule{.15em}{.05em}{.05em} 
 \multicolumn{2}{c|}{Fusion Method} & mAP & \multicolumn{20}{c}{AP per category} \\\specialrule{.15em}{.05em}{.05em} 
 \multicolumn{2}{c|}{Platt} & .695 & .734 & .761 & .758 & .580 & .574 & .720 & .820 & .761 & .538 & .677 & .685 & .749 & .769 & .716 & .781 & .454 & .685 & .677 & .748 & .715 \\
 \multicolumn{2}{c|}{WS} & .700 & .699 & .796 & .696 & .559 & .503 & .780 & .757 & .862 & .527 & .791 & .671 & .840 & .795 & .766 & .696 & .413 & .689 & .704 & .778 & .686 \\
 \multicolumn{2}{c|}{Bayes} & .706 & .740 & .786 & .783 & .607 & .525 & .775 & .816 & .854 & .480 & .763 & .652 & .840 & .810 & .738 & .759 & .448 & .700 & .565 & .760 & .746 \\
 \multicolumn{2}{c|}{LEF} & .727 & .733 & ..802 & .743 & .585 & .539 & .784 & .820 & {\bf .870} & .555 & .792 & .680 & .848 & .844 & .749 & .761 & .450 & .722 & .728 & .787 & .737 \\
 \multicolumn{2}{c|}{D2R} & .728 & .751 & .763 & .787 & .654 & .534 & .814 & .814 & .841 & .596 & .796 & .687 & .846 & .841 & .727 & .717 & .455 & .729 & .707 & .785 & .709 \\\hline
 \multicolumn{2}{c|}{DBF} & \underline{{\bf .760}} & \underline{{\bf .788}} & \underline{{\bf .807}} & \underline{{\bf .792}} & \underline{{\bf .661}} & \underline{{\bf .588}} & \underline{{\bf .848}} & \underline{{\bf .853}} & .860 & \underline{{\bf .599}} & {\bf .802} & \underline{{\bf .739}} & {\bf .862} & \underline{{\bf .859}} & \underline{{\bf .792}} & \underline{{\bf .786}} & \underline{{\bf .474}} & \underline{{\bf .753}} & \underline{{\bf .753}} & \underline{{\bf .808}} & \underline{{\bf .767}} \\\specialrule{.15em}{.05em}{.05em} 
\end{tabular}}
\label{tab:map_cnn_voc07}
\end{table*}

\subsubsection{Individual Detectors}

For the second evaluation, we use eight object detectors to verify effectiveness of the proposed method with various detectors that provide complementary features. Four of them are from the detector pool in the previous evaluation ({\bf DSIFT}, {\bf TAG}, {\bf ESVM}, and {\bf DPM}) and the rest are HOG~\cite{NDalalCVPR05}, DPM with color attribute features~\cite{FSKhanCVPR12}, and two CNN-based methods (Loc-CNN~\cite{MOquabCVPR14} and R-CNN~\cite{RGirshickPAMI16}). This detector pool leverages various image features including image gradient (SIFT, HOG), color attributes, and CNN features, and various classifiers including SVM and CNN.\medskip

\noindent{\bf HOG.} HOG belongs to a set of gradient-based features, such as SIFT~\cite{DLoweIJCV04}. However, compared to other gradient-based features, HOG is computed on a dense grid of uniformly spaced cells and followed by overlapping local contrast normalization. HOG features are used to represent object appearance. SVM is then trained to distinguish objects of interest from background.\medskip

\noindent{\bf Color Attributes.} Khan et al.~\cite{FSKhanCVPR12} use the DPM by replacing HOG with color attributes. Color attributes are compact, computationally efficient, and when combined with traditional shape features provide enhanced results for object detection.\medskip

\noindent{\bf Loc-CNN.} Loc-CNN is based on AlexNet~\cite{AKrizhevskyNIPS12} pre-trained on a very large image dataset, ImageNet~\cite{JDengCVPR09}. The target image dataset used in this evaluation (PASCAL VOC 07) contains much fewer images than ImageNet with quite different visual characteristics. To adapt the CNN structure of AlexNet to category distribution and characteristics of the target dataset, the final fully connected classification layer is learned again over the target dataset.

Note that Loc-CNN is designed for a task of object localization which aims to estimate approximate locations of objects-of-interest. Therefore, the detection results of Loc-CNN are not as accurate as those of state-of-the-art object detection methods being able to find tight bounding boxes containing the objects. These detection results of Loc-CNN, weak in providing accurate bounding boxes yet strong in finding objects, may provide useful information about rough locations of objects that can be used to enhance detection performance as a whole via fusion.\medskip

\noindent{\bf R-CNN.} R-CNN refers to a suite of CNN-based object detection methods that apply CNNs for bottom-up region proposals to localize objects. We use AlexNet~\cite{AKrizhevskyNIPS12} as a backbone of the R-CNN. In the following evaluation, we will consider higher-capacity CNNs, such as VGG-16~\cite{KSimonyanICLR15} and ResNet-101~\cite{KHeCVPR16} and enhanced R-CNNs, such as fast/faster R-CNN~\cite{RGirshickICCV15,SRenPAMI17}.

\subsubsection{Baseline Fusion Methods}

As a baseline, we add two more fusion methods to the three methods used in the previous evaluation: Local Expert Forest (LEF)~\cite{JLiuECCV12} and Detect2Rank (D2R)~\cite{SKaraogluTIP16}.\medskip

\noindent{\bf LEF.} Local expert forest employs a mixture of multiple experts, which provides binary output. For training each expert, training data is divided into two sets by $k$-means clustering ($k$=2) with random initialization. The expert is trained for assigning test data into one of the two sets. Multiple experts have different hyper-planes in the score space defined by the outputs of multiple classifiers according to partitions. Note that, for each partition, $k$-mean clustering is separately applied to positive and negative examples to cope with data imbalance issue. Each expert is trained by minimizing mean-square-error (MMSE).\medskip

\noindent{\bf D2R.} Detect2Rank adopts three context features from detection scores: detector-detector context, object-object relation, and object-saliency. First, the detector-detector context measures detection consistency among multiple detectors. Second, object-object relation calculates co-occurence between different objects. Lastly, object-saliency indicates how likely each detection contains an object of interest. A fusion score is computed via weighted sum over these three features. The weight is trained via ranking optimization.

\subsubsection{Detection Accuracy}

Detection accuracy of each individual detector and fusion methods are reported in Table~\ref{tab:map_pascal}. DBF provides the best detection accuracy in terms of mAP. DBF demonstrates the best results for 12 of 20 categories in the PASCAL VOC 07 dataset. Notably, DBF is the only fusion approach that outperforms R-CNN on the dataset though improvement is small. Only marginal improvement is achieved because the performance of R-CNN is much better than the other detectors. 

Therefore, we evaluated fusion performance again, but without R-CNN, and the results are presented in Table~\ref{tab:map_pascal2}. In Table~\ref{tab:map_pascal2}, DBF still outperformed all baseline fusion methods and all individual detectors by a considerably large margin (.06 from {\bf LEF} and .10 from {\bf Color attr.}). The fact that DBF outperforms Bayesian fusion demonstrates the benefits of incorporating an intermediate state into the set of hypotheses.

\begin{table*}[t]
\caption{{\bf Average Precision on the PASCAL VOC 12 dataset.}}
\begin{subfigure}{\linewidth}
\setlength{\tabcolsep}{2.6pt}
\renewcommand{\arraystretch}{1.4}
\centering
{\scriptsize
\begin{tabular}{ll|c|cccccccccccccccccccc}
\specialrule{.15em}{.05em}{.05em} 
\multicolumn{2}{c|}{Backbone} & \multirow{2}{2.5em}{mAP} & \multicolumn{20}{c}{AP per category} \\
img. classif. & obj. detect & & aero & bike & bird & boat & bottle & bus & car & cat & chair & cow & table & dog & horse & mbike & persn & plant & sheep & sofa & train & tv 
\\\specialrule{.15em}{.05em}{.05em} 
\multirow{2}{5em}{VGG M} & Fast R-CNN\textsuperscript{1} & .580 &  .768 & .671 & .571 & .363 & .280 & .670 & .600 & .812 & .313 & .613 & .463 & .791 & .712 & .731 & .625 & .262 & .598 & .496 & .678 & .580 \\
 & Faster R-CNN\textsuperscript{2} & .568 & .746 & .651 & .524 & .339 & .342 & .653 & .618 & .793 & .304 & .592 & .412 & .773 & .693 & .705 & .687 & .274 & .610 & .443 & .639 & .555 \\\hline
\multirow{2}{5em}{VGG 16} & Fast R-CNN\textsuperscript{3} & .666 & .824 & .763 & .688 & .482 & .376 & .746 & .691 & .881 & .416 & .726 & .522 & .861 & .792 & .798 & .726 & .359 & .684 & .614 & .748 & .614 \\
 & Faster R-CNN\textsuperscript{4} & .673 & .824 & .747 & .704 & .489 & .497 & .732 & .739 & .866 & .442 & .745 & .485 & .860 & .788 & .763 & .785 & .390 & .675 & .566 & .749 & .621 \\\hline
\multirow{2}{5em}{ResNet 101} & Fast R-CNN\textsuperscript{5} & .687 & .830 & .781 & .722 & .513 & .390 & .756 & .703 & .915 & .434 & .752 & .549 & .894 & .825 & .790 & .739 & .354 & .711 & {\bf .635} & {\bf .809} & .636 \\
 & Faster R-CNN\textsuperscript{6} & {\bf .727} & {\bf .853} & {\bf .792} & \underline{{\bf .774}} & {\bf .561} & {\bf .548} & {\bf .780} & {\bf .755} & \underline{{\bf .918}} & {\bf .501} & \underline{{\bf .800}} & {\bf .554} & {\bf .906} & {\bf .837} & {\bf .824} & \underline{{\bf .803}} & {\bf .471} & {\bf .758} & .616 & {\bf .809} & {\bf .680} \\\specialrule{.15em}{.05em}{.05em} \specialrule{.15em}{.05em}{.05em} 
 \multicolumn{2}{c|}{Fusion Method} & mAP & \multicolumn{20}{c}{AP per category} \\\specialrule{.15em}{.05em}{.05em} 
 \multicolumn{2}{c|}{Platt\textsuperscript{7}} & .664 & .836 & .722 & .750 & .503 & .549 & .736 & .751 & .752 & .477 & .692 & .517 & .796 & .609 & .767 & .769 & .465 & .661 & .476 & .799 & .663 \\
 \multicolumn{2}{c|}{WS\textsuperscript{8}} & .682 & .828 & .770 & .713 & .519 & .483 & .736 & .718 & .881 & .460 & .762 & .520 & .863 & .786 & .782 & .678 & .425 & .711 & .606 & .776 & .613 \\
 \multicolumn{2}{c|}{Bayes\textsuperscript{9}} & .688 & .847 & .771 & .730 & .521 & .510 & .748 & .741 & .891 & .462 & .736 & .497 & .881 & .795 & .798 & .766 & .433 & .704 & .473 & .778 & .679 \\
 \multicolumn{2}{c|}{LEF\textsuperscript{10}} & .717 & .845 & .793 & .734 & .541 & .520 & .766 & .754 & .907 & .505 & .797 & .565 & .886 & .815 & .812 & .767 & .443 & .746 & .654 & .800 & .689 \\
 \multicolumn{2}{c|}{D2R\textsuperscript{11}} & .714 & .852 & .794 & .769 & .553 & .520 & .783 & .743 & .901 & .484 & .778 & .533 & .893 & .831 & .814 & .773 & .443 & .737 & .629 & .796 & .646 \\\hline
 \multicolumn{2}{c|}{DBF\textsuperscript{12}} & \underline{{\bf .739}} & \underline{{\bf .860}} & \underline{{\bf .813}} & \underline{{\bf .774}} & \underline{{\bf .600}} & \underline{{\bf .557}} & \underline{{\bf .789}} & \underline{{\bf .775}} & {\bf .916} & \underline{{\bf .521}} & {\bf .799} & \underline{{\bf .581}} & \underline{{\bf .910}} & \underline{{\bf .845}} & \underline{{\bf .826}} & {\bf .802} & \underline{{\bf .478}} & \underline{{\bf .759}} & \underline{{\bf .669}} & \underline{{\bf .819}} & \underline{{\bf .694}} \\\specialrule{.15em}{.05em}{.05em}
\end{tabular}}
\end{subfigure}

\begin{subfigure}{\linewidth}
\setlength{\tabcolsep}{15.0pt}
\centering
{\tiny
\begin{tabular}{lll}
\textsuperscript{1}\url{http://host.robots.ox.ac.uk:8080/anonymous/MWQ2S0.html} &
\textsuperscript{2}\url{http://host.robots.ox.ac.uk:8080/anonymous/ZNYABA.html} &
\textsuperscript{3}\url{http://host.robots.ox.ac.uk:8080/anonymous/6GW5CG.html} \\
\textsuperscript{4}\url{http://host.robots.ox.ac.uk:8080/anonymous/II2XAA.html} &
\textsuperscript{5}\url{http://host.robots.ox.ac.uk:8080/anonymous/9NFQNR.html} &
\textsuperscript{6}\url{http://host.robots.ox.ac.uk:8080/anonymous/T57ZJT.html} \\
\textsuperscript{7}\url{http://host.robots.ox.ac.uk:8080/anonymous/5IDSL1.html} &
\textsuperscript{8}\url{http://host.robots.ox.ac.uk:8080/anonymous/JMDPTJ.html} &
\textsuperscript{9}\url{http://host.robots.ox.ac.uk:8080/anonymous/LHYOXS.html} \\
\textsuperscript{10}\url{http://host.robots.ox.ac.uk:8080/anonymous/DD20VJ.html} &
\textsuperscript{11}\url{http://host.robots.ox.ac.uk:8080/anonymous/T8XIJR.html} &
\textsuperscript{12}\url{http://host.robots.ox.ac.uk:8080/anonymous/M54FPE.html} \\
\end{tabular}
}
\end{subfigure}
\label{tab:map_cnn_voc12}
\end{table*}

\subsection{Evaluation of PASCAL VOC 07 \& 12 with R-CNNs}

\subsubsection{Individual Detectors}

In this evaluation, we only use CNN-based object detectors as individual detectors. Since R-CNN~\cite{RGirshickPAMI16} was introduced, its descendents~\cite{RGirshickICCV15,SRenPAMI17,JDaiNIPS16,KHeICCV17,HLeeArxiv17} have shown remarkable object detection performance with respect to detection accuracy as well as much faster running time. Among various R-CNNs, we choose two methods: Fast R-CNN~\cite{RGirshickICCV15} and Faster R-CNN~\cite{SRenPAMI17}. 

Detection accuracy of R-CNN highly depends on a backbone CNN architecture trained on a large scale ImageNet dataset. Among many CNNs, we consider three architectures: VGG-M~\cite{KChatfieldBMVC14}, VGG-16~\cite{KSimonyanICLR15}, and ResNet-101~\cite{KHeCVPR16}. As individual detectors, we use six CNN-based methods which compose all combinations of two R-CNNs and three backbone CNNs.\medskip

\noindent{\bf Fast R-CNN.} While R-CNN consists of multiple pipelines which lead to slow computation, Fast R-CNN is built as unified architecture beside a region proposal generation and is trained in an end-to-end fashion. This innovation improves training and test speed while also increasing detection accuracy.\medskip

\noindent{\bf Faster R-CNN.} In Fast R-CNN, region proposal is a major bottleneck in computation. To improve speed, Faster R-CNN incorporates the cost-free region proposal network (RPN) into the Fast R-CNN architecture by training RPN with the entire network in an end-to-end fashion, resulting in significant increase in speed.\medskip

\noindent{\bf VGG-M.} The VGG-M network consists of five convolutional layers and three fully connected layers. This network has the same depth as AlexNet~\cite{AKrizhevskyNIPS12} but its width is larger than that of AlexNet.\medskip

\noindent{\bf VGG-16.} VGG-16 shares the similar architecture as VGG-M but is deeper than VGG-M by expanding the depth of convolutional layers of the VGG-M. While VGG-M has one layer in each convolutional module between two pooling layers, the VGG-16 network has two or three convolutional layers. Consequently, VGG-16 has thirteen convolutional layers and three fully connected layers.\medskip

\noindent{\bf ResNet-101.} ResNet-101 uses residual modules, which are designed to learn with reference to the residual input, the difference between the original input and the desired output. ResNet-101, substantially deeper than other networks (101 vs 16), is easily learned by incorporating these residual modules into the architecture.

\begin{figure*}[t]
    \centering
    	\includegraphics[trim = 0mm 0mm 0mm 0mm,clip=true,width=0.45\textwidth]{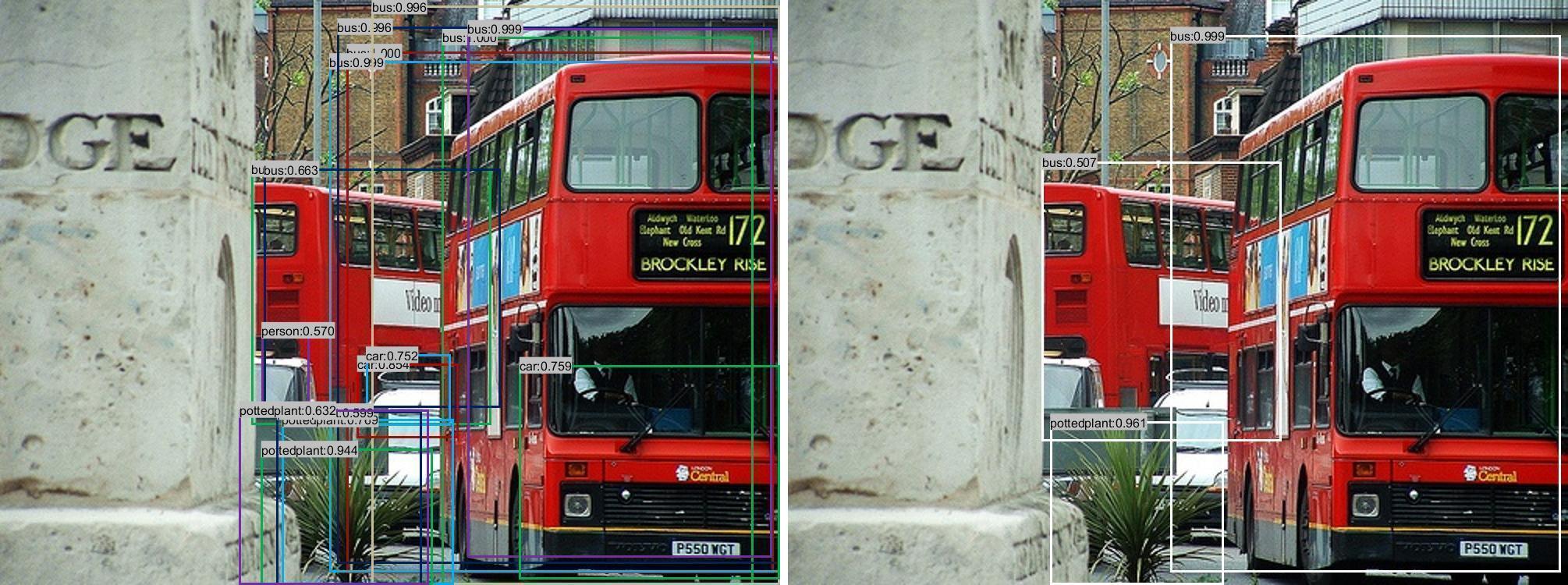}~~
    	\includegraphics[trim = 0mm 0mm 0mm 0mm,clip=true,width=0.45\textwidth]{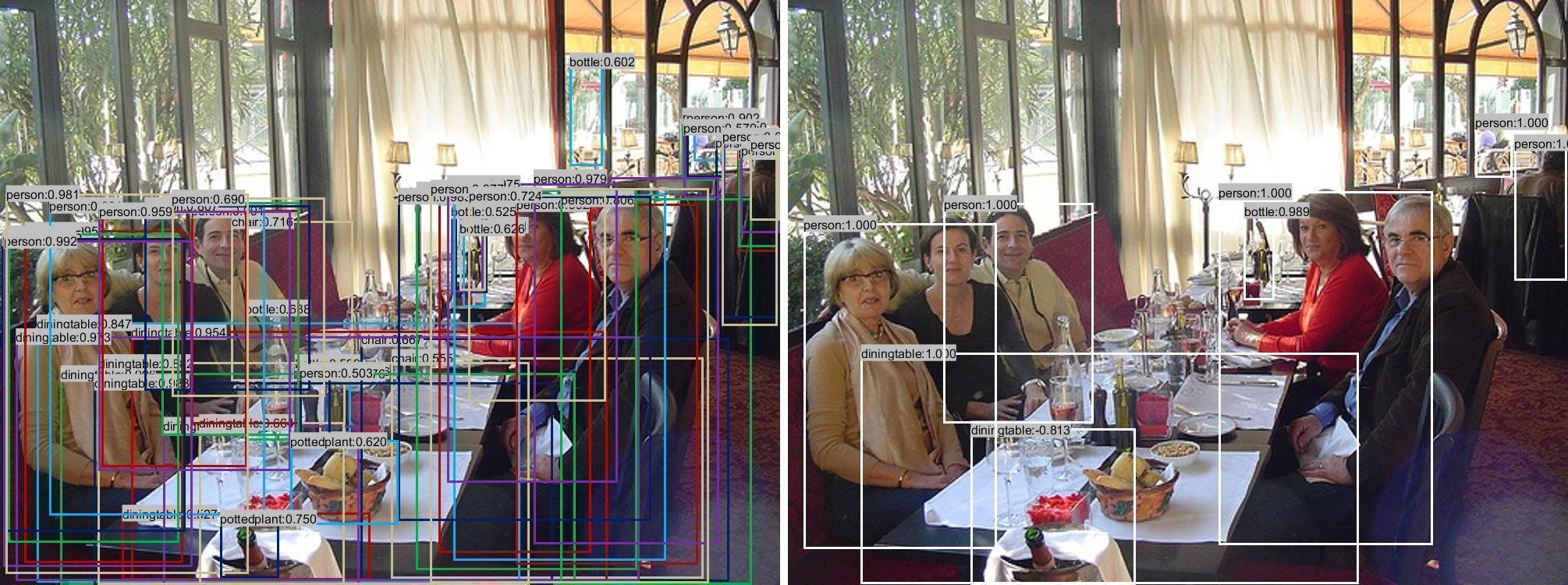}\\\vspace{0.7em}
    	\includegraphics[trim = 0mm 0mm 0mm 0mm,clip=true,width=0.45\textwidth]{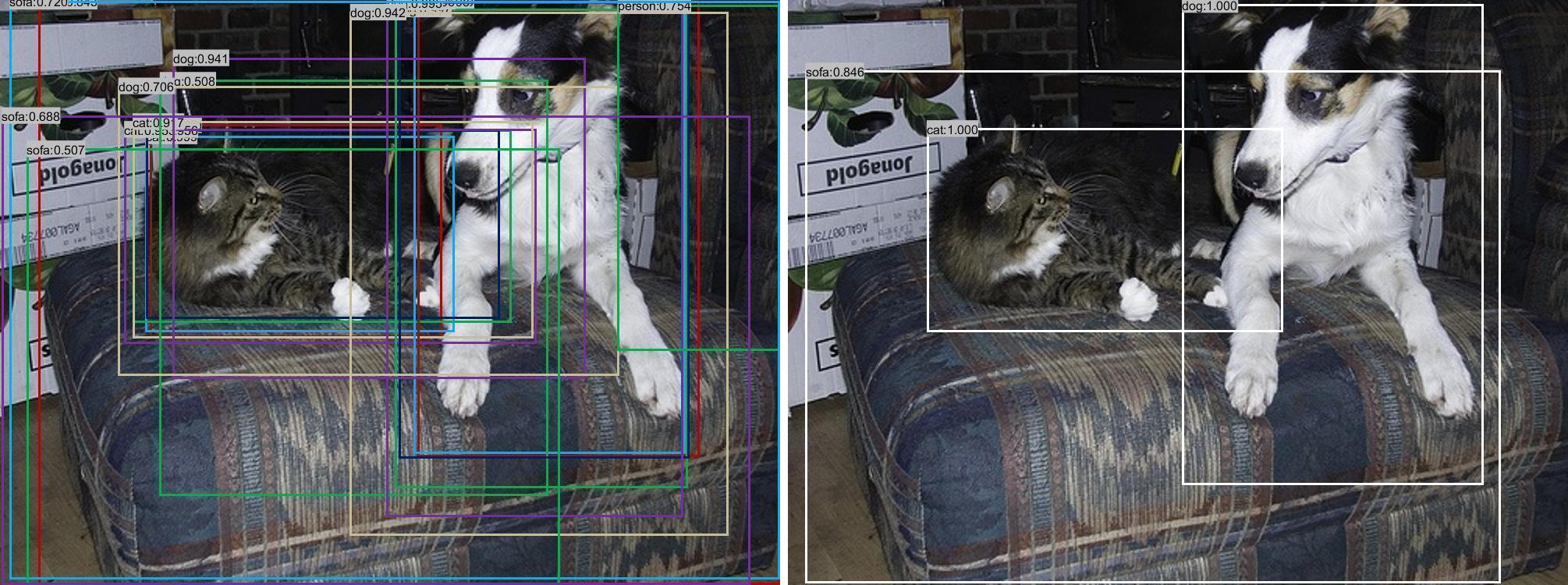}~~
    	\includegraphics[trim = 0mm 0mm 0mm 0mm,clip=true,width=0.45\textwidth]{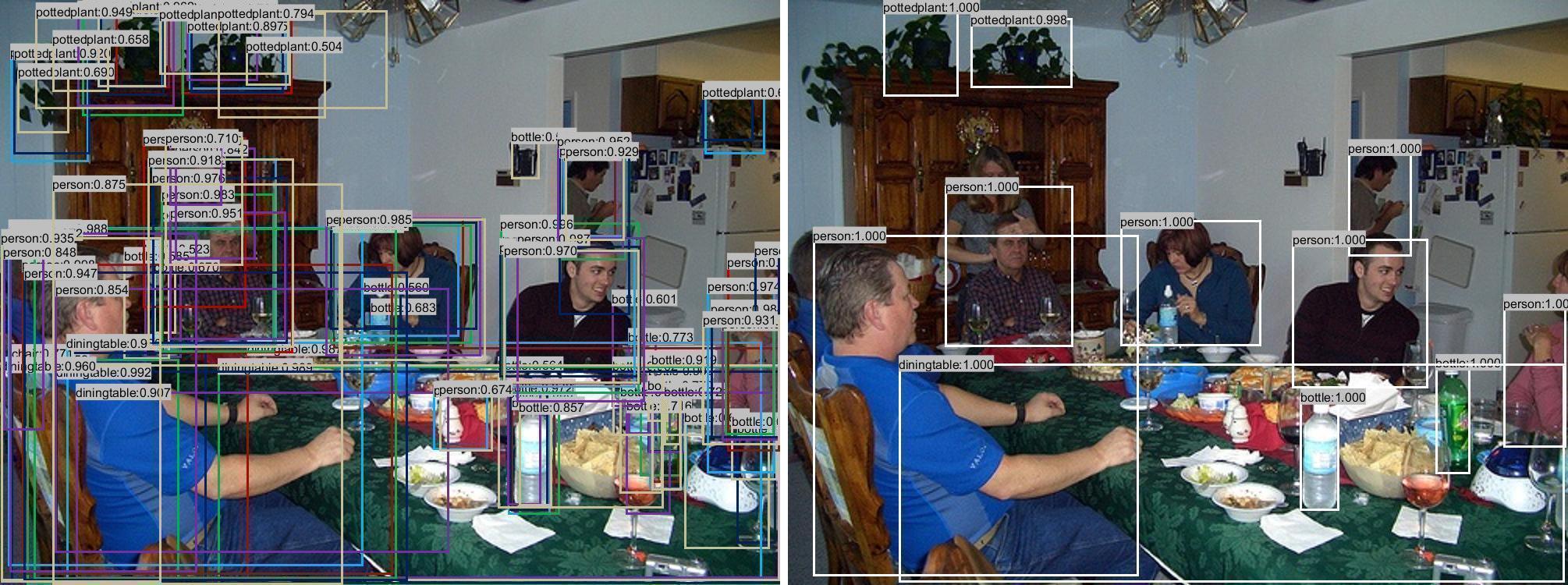}\\\vspace{0.7em}
    	\includegraphics[trim = 0mm 0mm 0mm 0mm,clip=true,width=0.45\textwidth]{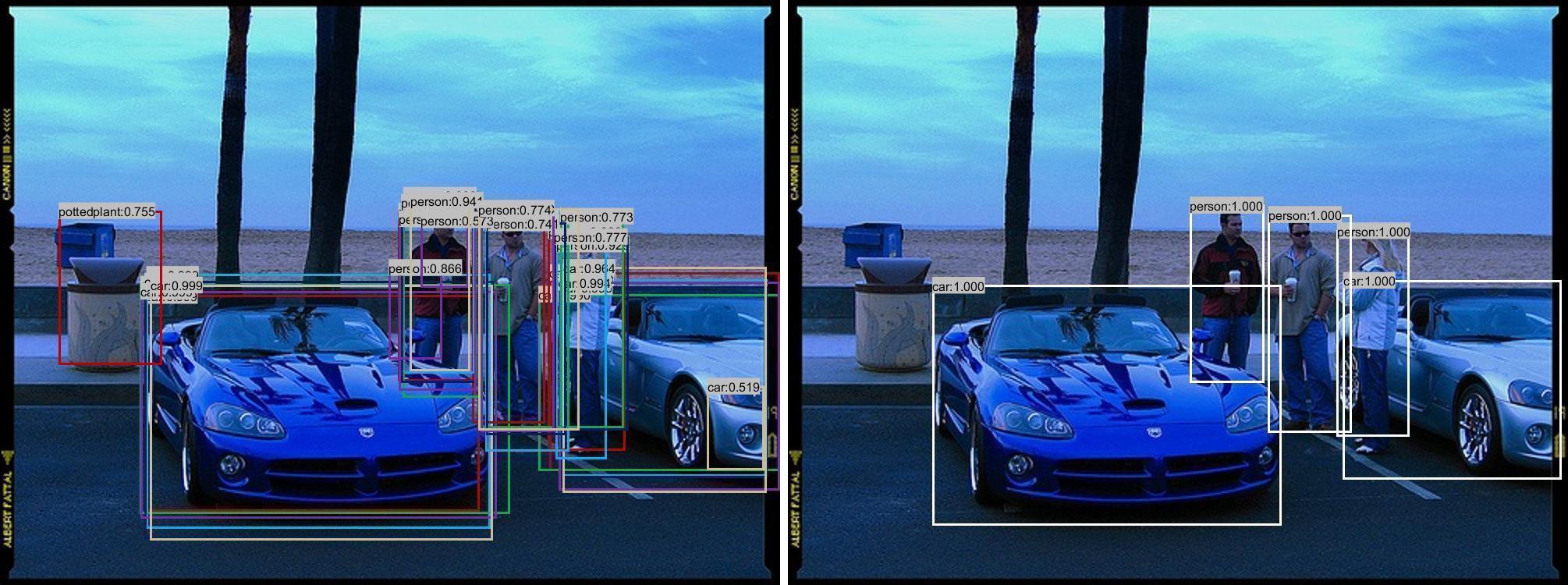}~~
    	\includegraphics[trim = 0mm 0mm 0mm 0mm,clip=true,width=0.45\textwidth]{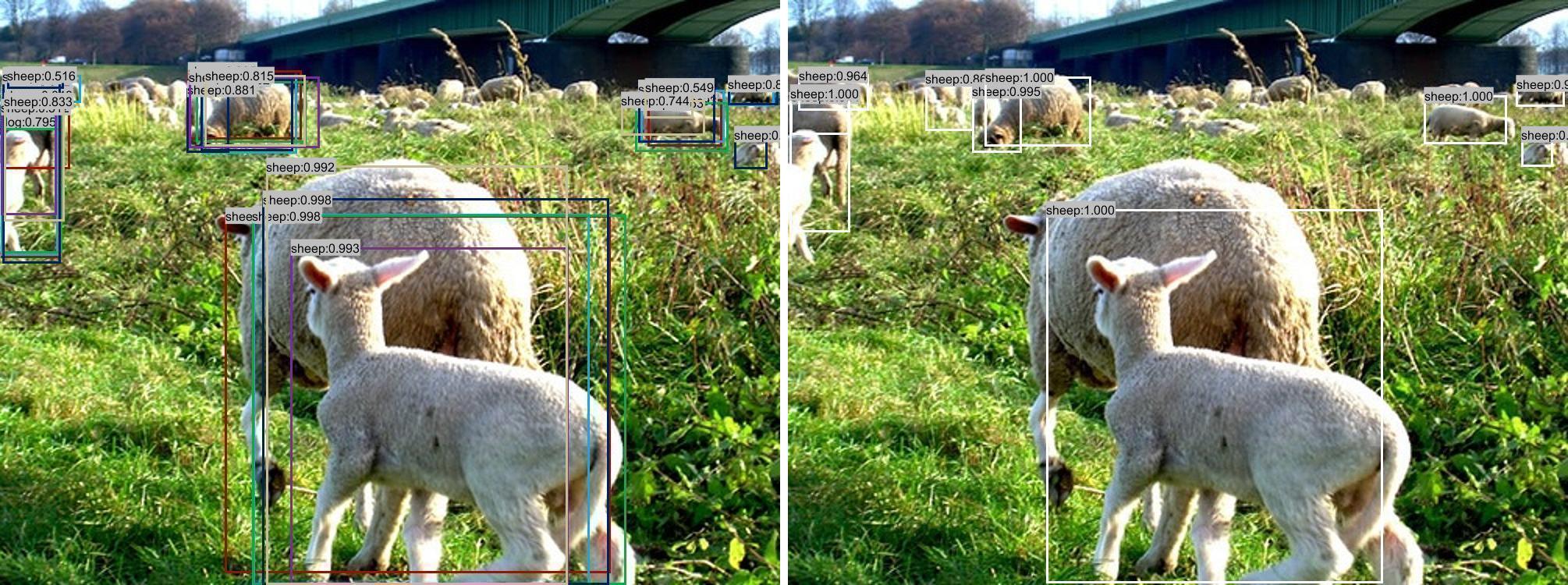}\\\vspace{0.7em}
    	\includegraphics[trim = 0mm 0mm 0mm 0mm,clip=true,width=0.45\textwidth]{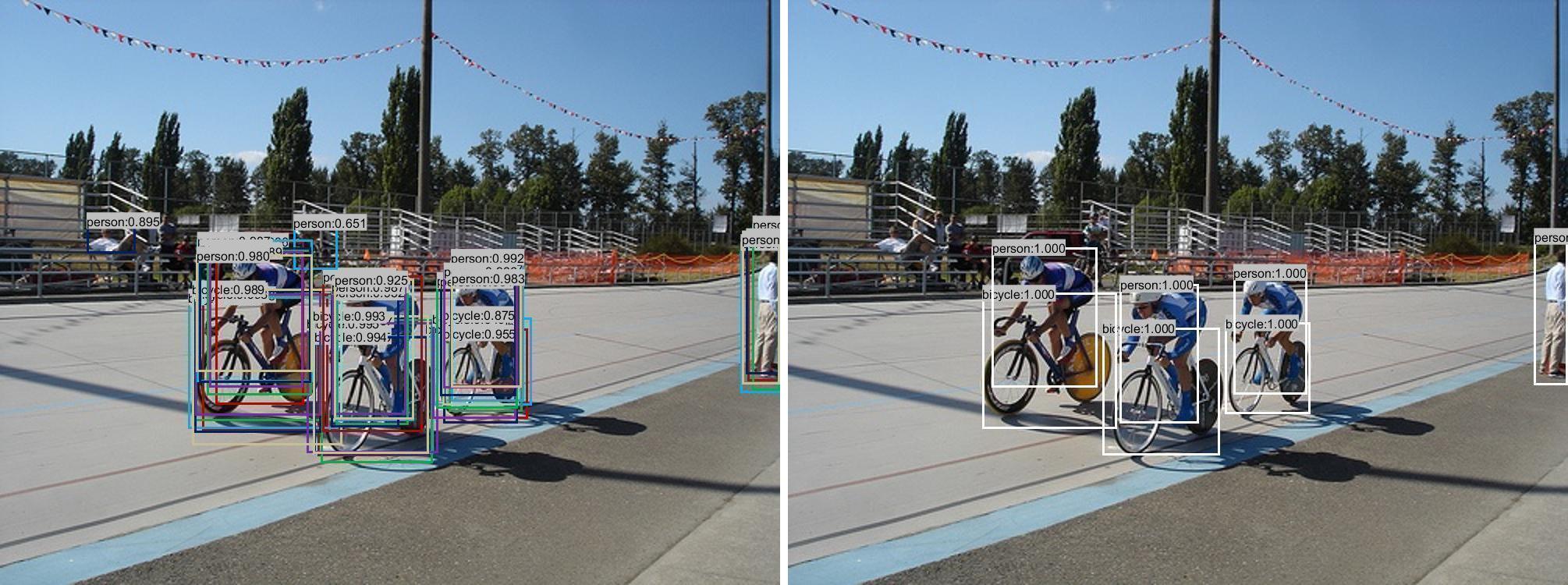}~~
    	\includegraphics[trim = 0mm 0mm 0mm 0mm,clip=true,width=0.45\textwidth]{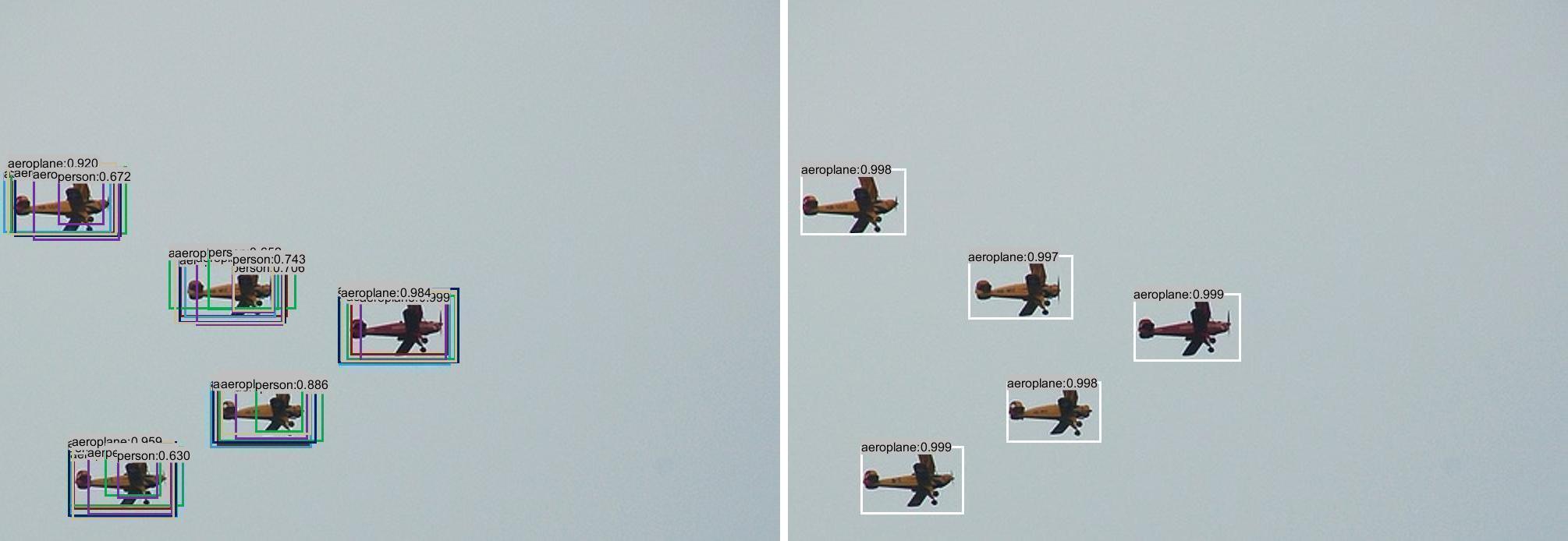}\\\vspace{0.7em}
    	\includegraphics[trim = 0mm 0mm 0mm 0mm,clip=true,width=0.29\textwidth]{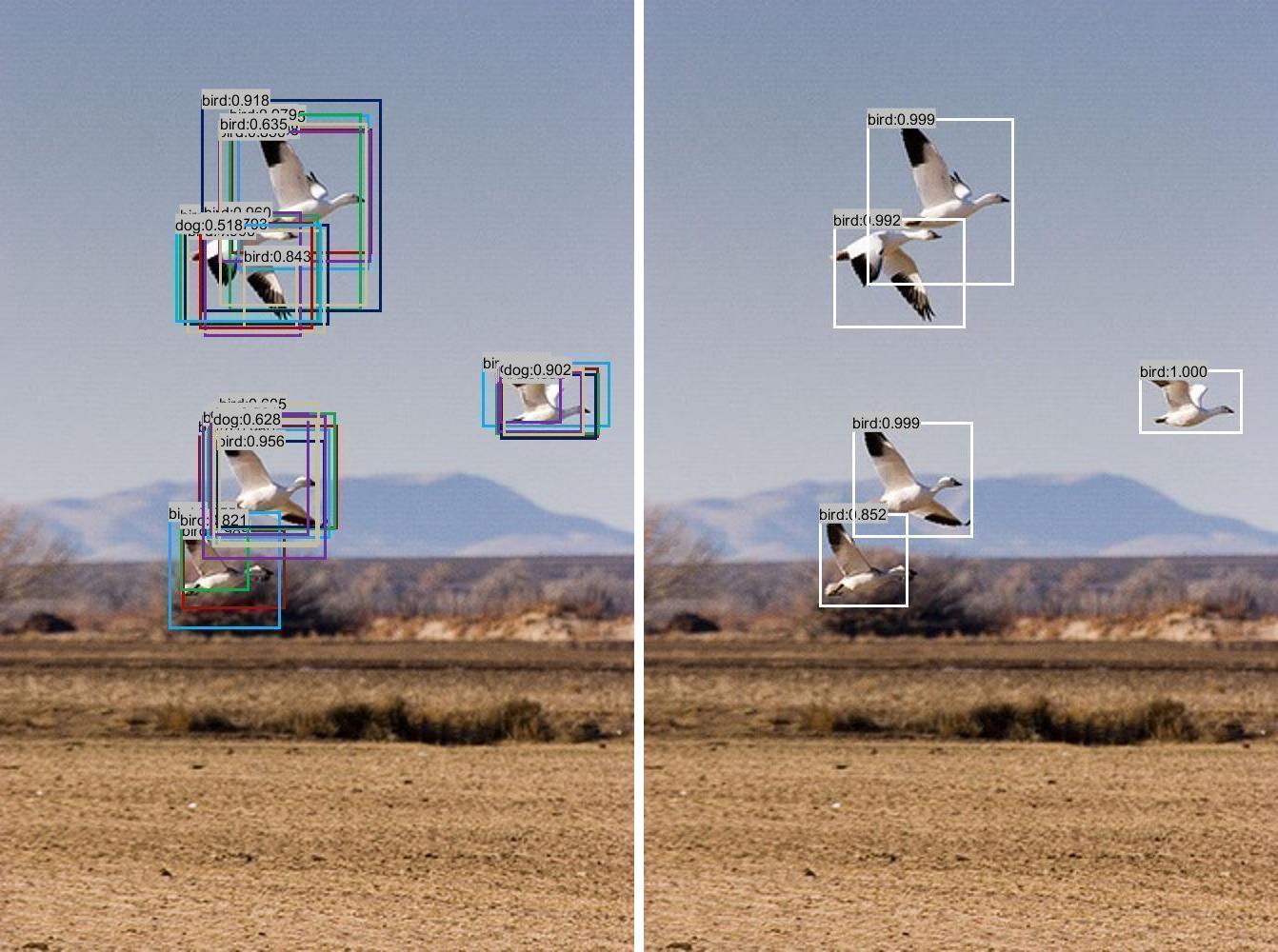}~~
    	\includegraphics[trim = 0mm 0mm 0mm 0mm,clip=true,width=0.383\textwidth]{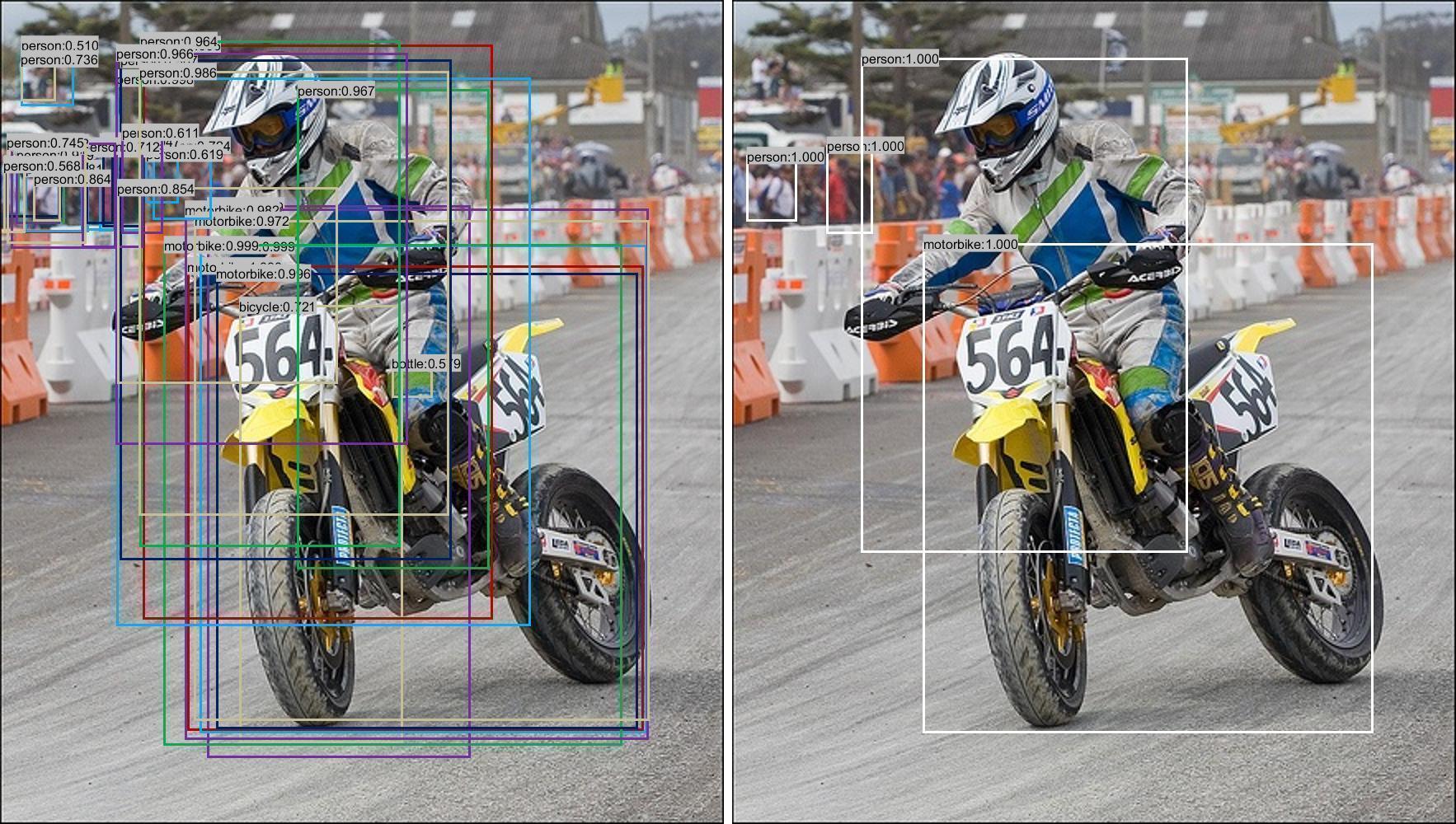}~~
    	\includegraphics[trim = 0mm 0mm 0mm 0mm,clip=true,width=0.29\textwidth]{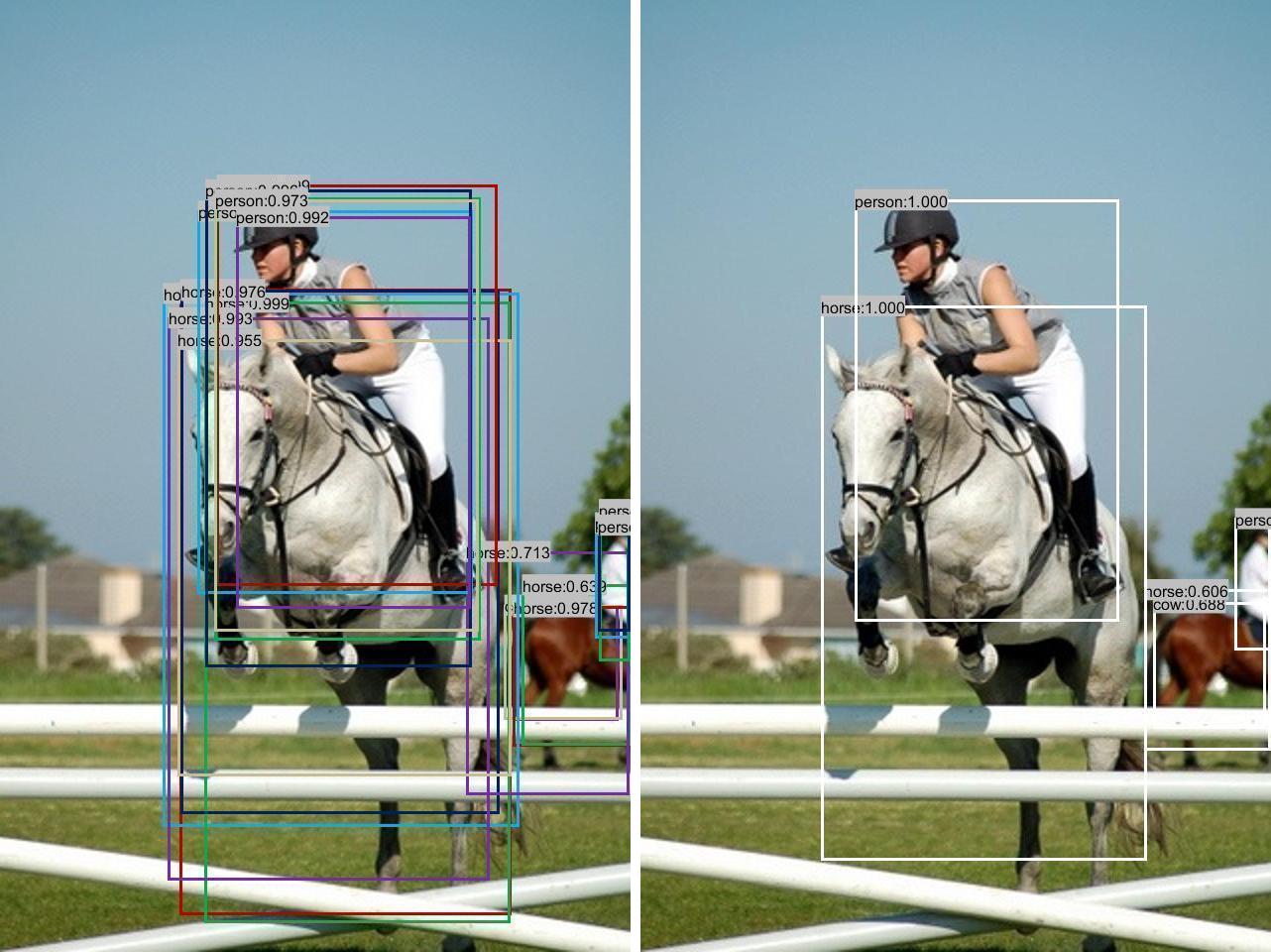}
    \caption{{\bf Example detections of R-CNNs and DBF.} For each pair of images, the left and right images show detections from six different R-CNNs and DBF, respectively. For R-CNNs, bounding boxes colored by    \textcolor{myfrrcn_l}{red}, \textcolor{myfrcn_l}{blue}, \textcolor{myfrrcn_m}{green}, \textcolor{myfrcn_m}{navy}, \textcolor{myfrrcn_s}{purple}, and \textcolor{myfrcn_s}{ivory} indicate detections of Faster R-CNN+ResNet-101, Fast R-CNN+ResNet-101, Faster R-CNN+VGG 16, Fast R-CNN+VGG 16, Faster R-CNN+VGG M, and Fast R-CNN+VGG M, respectively.}
    \label{fig:dbf_exmples}
\end{figure*}

\subsubsection{Baseline Fusion Methods}

We use the same set of baseline methods as the previous evaluation: {\bf Platt}, {\bf WS}, {\bf Bayes}, {\bf LEF}, and {\bf D2R}.

\subsubsection{Detection Accuracy}

On both PASCAL VOC 07 and 12 datasets as shown in Tables~\ref{tab:map_cnn_voc07} and~\ref{tab:map_cnn_voc12}, DBF presents the best detection accuracy among methods including all individual detectors and all fusion methods. On VOC 07, DBF provides enhanced detection accuracy over the best individual detector (the Faster R-CNN with ResNet-101) by .021 point in mAP while outperforming other fusion approaches by .32 point in mAP compared to the second best fusion approach, D2R. On VOC 12, DBF provides enhanced accuracy over the best individual detector as well as the second best fusion approach by .012 and .022 points in mAP, respectively. As shown in Tables~\ref{tab:map_cnn_voc07} and~\ref{tab:map_cnn_voc12}, DBF is the only fusion method, which provided higher accuracy than that of the best individual detector. Figure~\ref{fig:dbf_exmples} illustrates several examples of fusion results obtained by using DBF.

\section{Ablation Studies}

We have carried out in-depth ablation studies to demonstrate the strength of the proposed fusion approach, DBF.

\subsection{How Effective Is Dynamic Basic Probability Assignment?}

\begin{figure*}[t]
    \centering
    \includegraphics[trim = 85mm 25mm 70mm 0mm,clip=true,width=\linewidth]{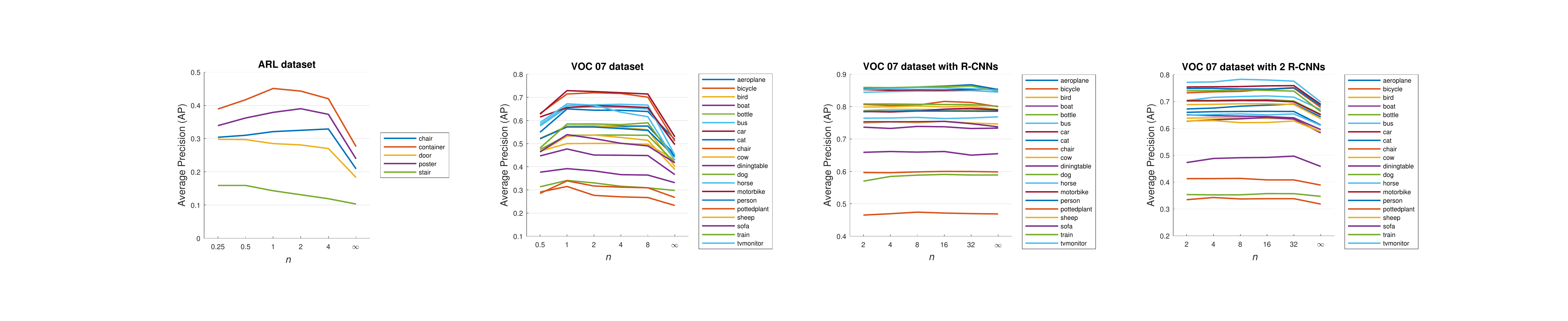}
    \caption{{\bf Comparison of fusion performance with respect to the various theoretical detectors.} $n$ in $x$ axis is the exponent in Eq.~\ref{eq:fn_bpd}. Three left plots shows the comparisons in three evaluation settings. The right most plot shows performance variation in mAP with respect to the parameter $n$ when using two weak R-CNNs (Fast R-CNN+VGG M and Faster R-CNN+VGG M) for fusion.}
    \label{fig:param_n}
\end{figure*}

\begin{figure*}[t]
    \centering
        \subfloat[Fast R-CNN + VGG M] {
    	\includegraphics[trim = 0mm 100mm 0mm 105mm,clip=true,width=0.33\textwidth]{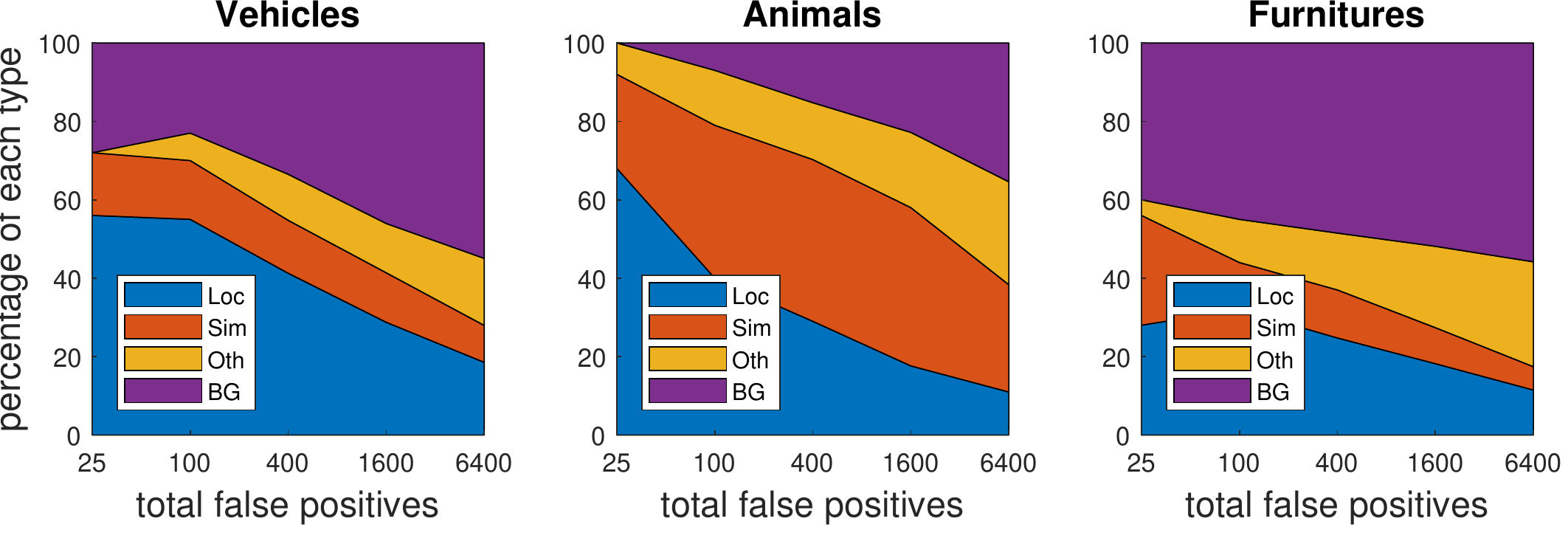}}
        \subfloat[Fast R-CNN + VGG 16] {
    	\includegraphics[trim = 0mm 100mm 0mm 105mm,clip=true,width=0.33\textwidth]{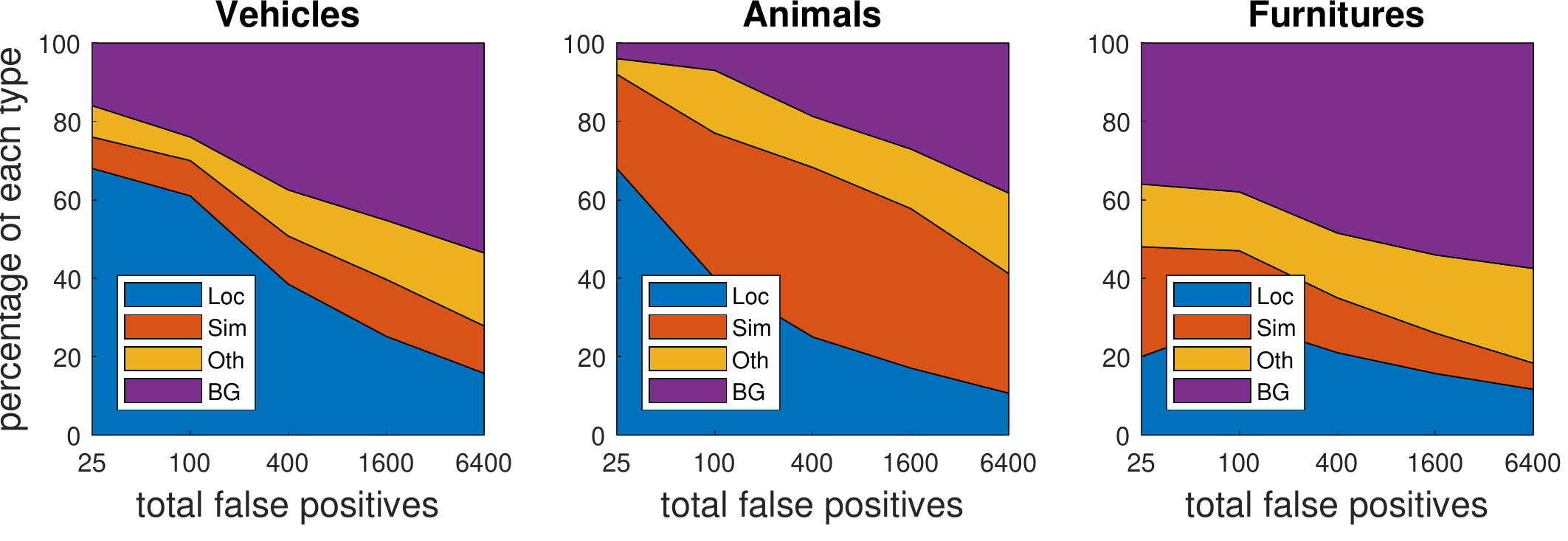}}
        \subfloat[Fast R-CNN + ResNet 101] {
    	\includegraphics[trim = 0mm 100mm 0mm 105mm,clip=true,width=0.33\textwidth]{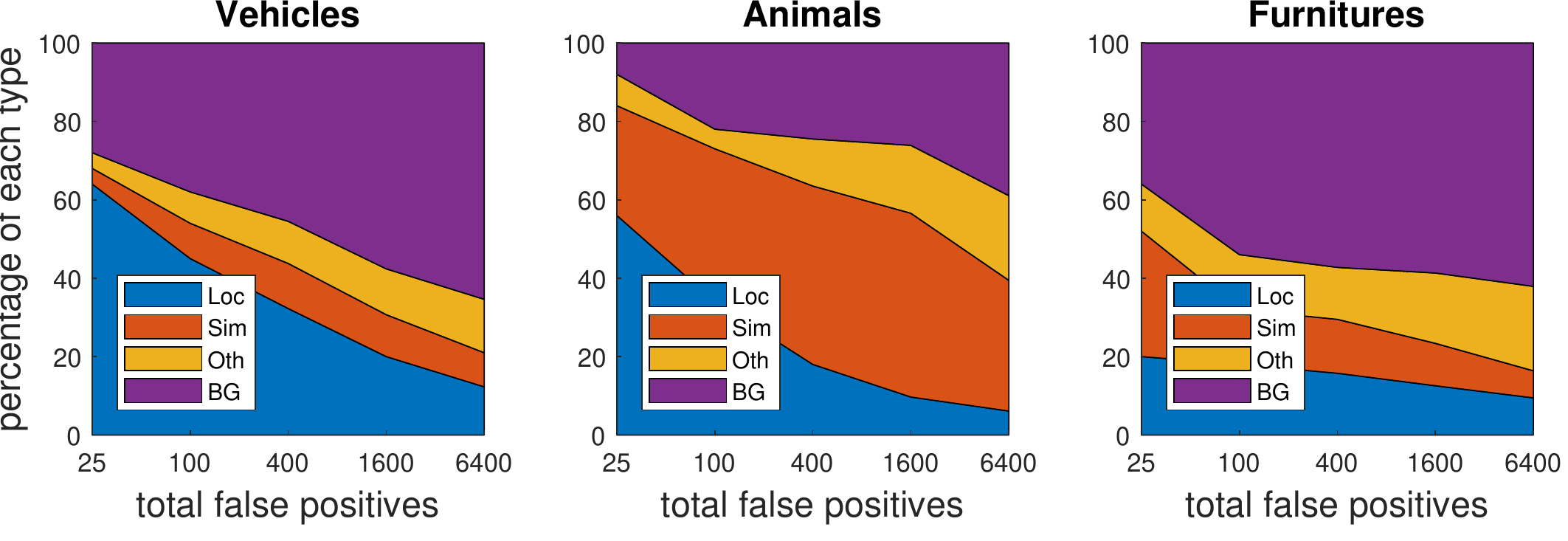}} \\
        \subfloat[Faster R-CNN + VGG M] {
    	\includegraphics[trim = 0mm 100mm 0mm 105mm,clip=true,width=0.33\textwidth]{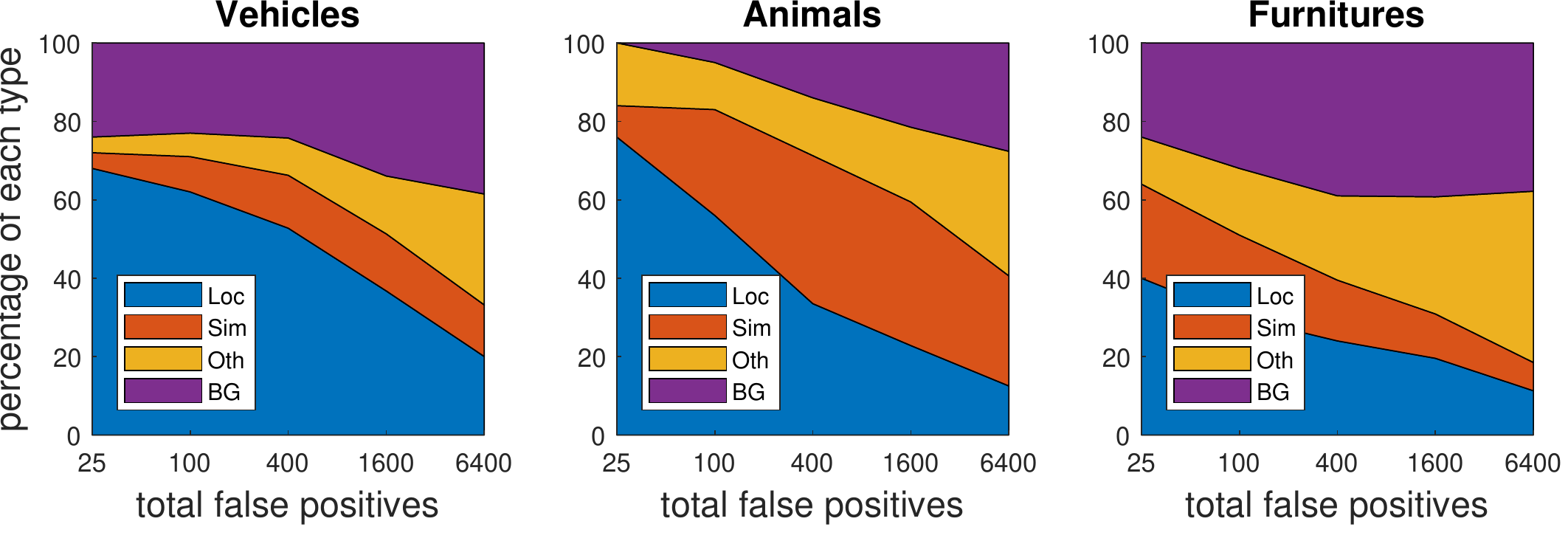}}
        \subfloat[Faster R-CNN + VGG 16] {
    	\includegraphics[trim = 0mm 100mm 0mm 105mm,clip=true,width=0.33\textwidth]{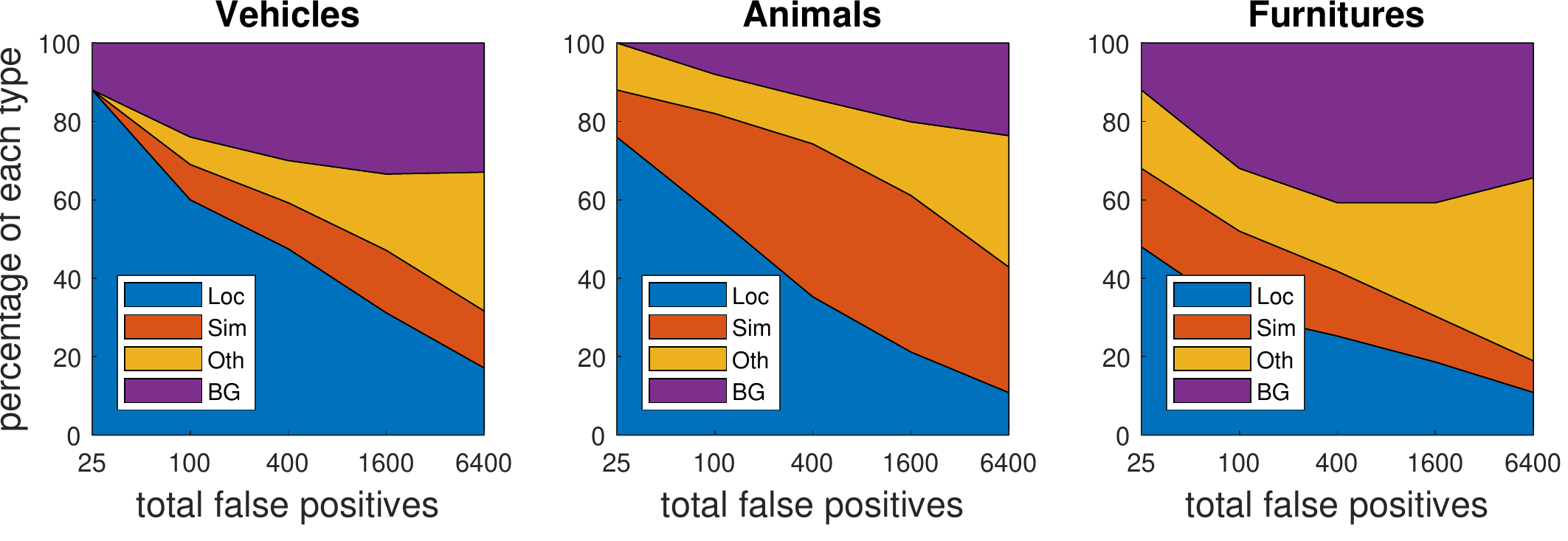}}
        \subfloat[Faster R-CNN + ResNet 101] {
    	\includegraphics[trim = 0mm 100mm 0mm 105mm,clip=true,width=0.33\textwidth]{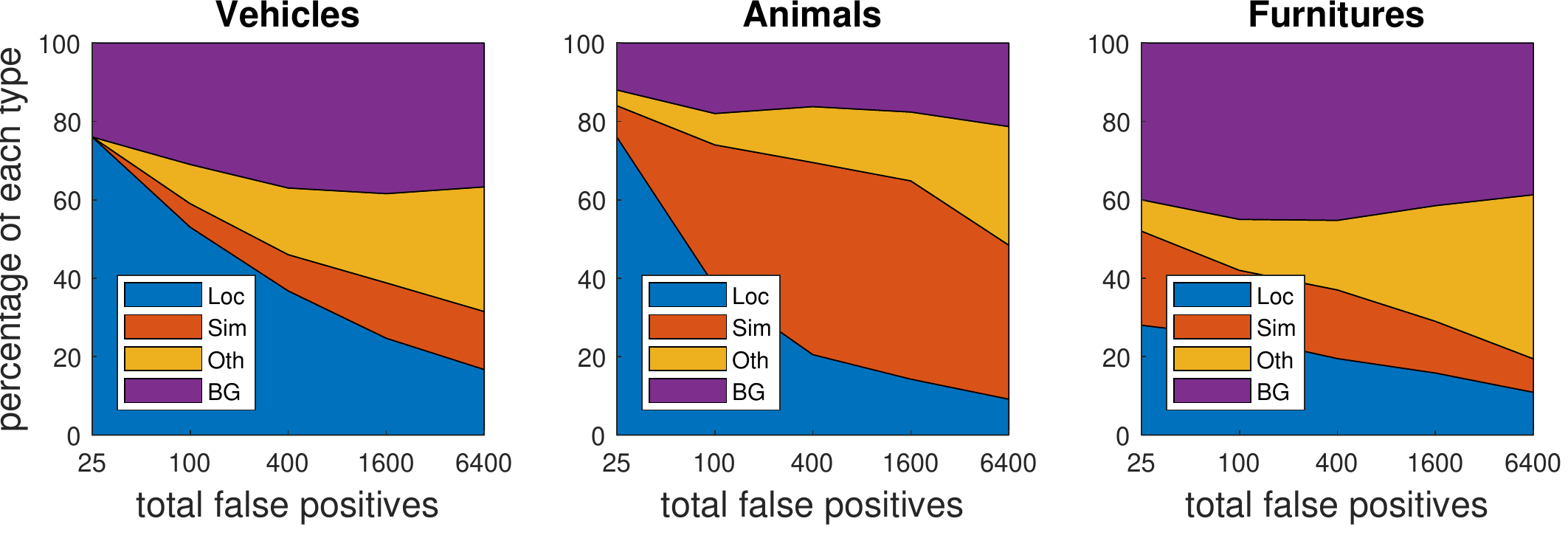}} \\
        \subfloat[DBF] {
    	\includegraphics[trim = 0mm 100mm 0mm 105mm,clip=true,width=0.33\textwidth]{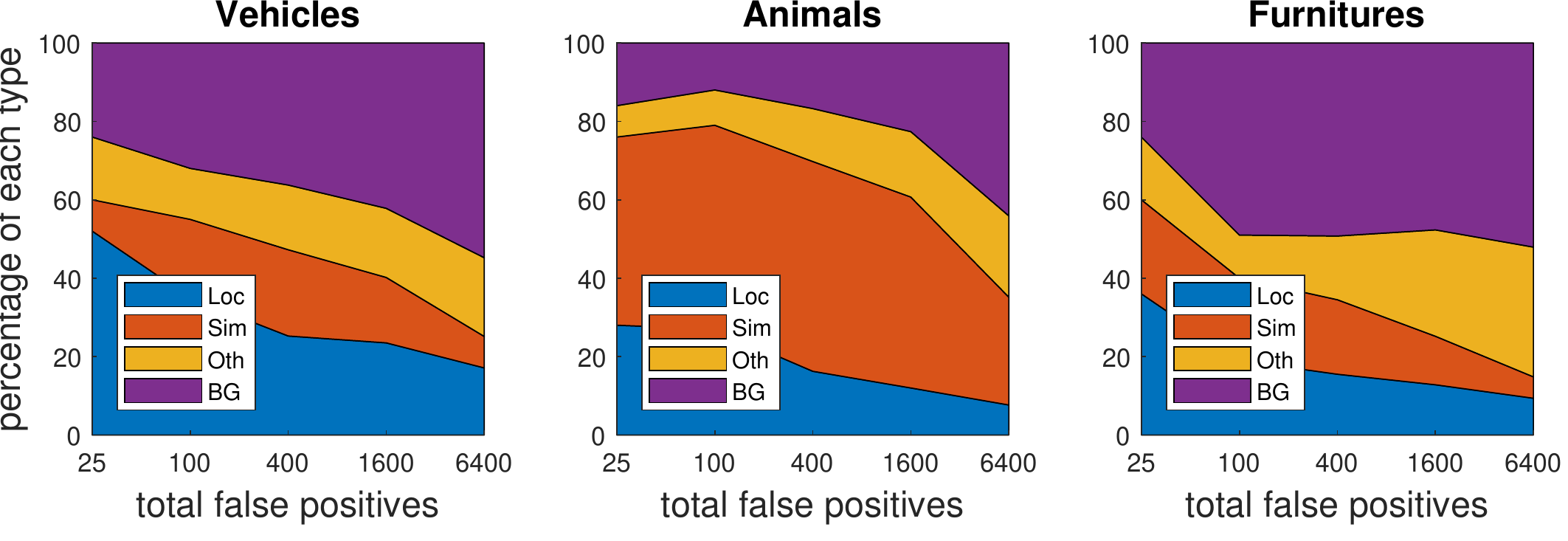}}
    \caption{{\bf Analysis of top-ranked false positives.} Each plot shows an evolving distribution of four FP types as detection scores are decreased.  The four types of false positives (FP) are 1) poor localization (Loc, a detection with an IoU overlap with the correct class between 0.2 and 0.5, or a duplicate), 2) confusion with similar classes (Sim), 3) Confusion with dissimilar object categories (Oth), and 4) confusion with background (BG). The analysis is performed on PASCAL VOC 07 dataset.  Among 20 object categories in PASCAL VOC 07 dataset, all vehicles belong to ``Vehicles'', all animals are in ``Animals'', and `chair', `diningtable', and `sofa' are assigned to ``Furniture''.}
    \label{fig:diag_fp}
\end{figure*}

\begin{table}[t]
\caption{{\bf Detection accuracy of DBF and Dempster-Shafer Theory (DST) in mAP.} Detection performance of DST-based fusion method depends on a precision value. Table shows detection accuracy of different DST-based fusion methods over various precisions.}
\setlength{\tabcolsep}{9.5pt}
\renewcommand{\arraystretch}{1.4}
\centering
\begin{tabular}{c|ccccc|c}
\specialrule{.15em}{.05em}{.05em}
 & \multicolumn{5}{c|}{DST} & DBF \\\specialrule{.15em}{.05em}{.05em}
precision & .4 & .5 & .6 & .7 & .8 & $\cdot$ \\\hline
mAP & .461 & .510 & .517 & .580 & .601 & .760 \\\specialrule{.15em}{.05em}{.05em}
\end{tabular}
\label{tab:dbf_vs_dst}
\end{table}

To demonstrate the advantages of the dynamic basic probability assignment which is the main component of DBF, we compare DBF to plain Dempster-Shafer Theory (DST) with respect to a detection accuracy. DST uses only static basic probability assignment~\cite{LXuSMC92}, in which each detector's prior performance is characterized by the probabilities of the three hypotheses at a certain point of precision. Once a detection score becomes larger than the threshold corresponding to the precision value, these probabilities are assigned to the hypotheses. Otherwise, zero probabilities are assigned to basic probability of $T$ and $\neg T$ while basic probability of $I$ becomes one. We calculate the detection accuracy of DST from multiple settings where a precision value is changed according to different recall values of 0.4, 0.5, $\cdots$, 0.8.

Table~\ref{tab:dbf_vs_dst} shows the detection accuracy obtained by DBF and DST with various precision values. In terms of mAP, DBF outperforms DST over any of the precision values by a significantly large margin, which strongly supports the effectiveness of dynamic basic probability assignment.

\subsection{Is the Theoretical Detector Necessary?}

Figure~\ref{fig:param_n} illustrates the variation in mAP for each object category in the datasets, as the shape of the PR curve of the theoretical detector changes. The optimal value of the parameter $n$ (the exponent in Equation~\ref{eq:fn_bpd}), which dictates the shape (and hence, estimated performance) of the theoretical detector, is different for different object categories. However, note that the ideal detector ($n = \infty$) underperforms DBF over other choices of $n$ with respect to every object category on the first two evaluations (left two plots in the Figure). This result suggests that our method of splitting the false positives into \emph{non-target} ($\neg T$) and \emph{intermediate} ($I$) hypotheses is actually beneficial. 

On the other hand, detection accuracy does not greatly change with respect to the parameter $n$ for the third evaluation.  The ideal detector ($n=\infty$) also provides similar detection accuracy for every object category. To further analyze this performance, we carried out fusion with two weak R-CNNs (Fast R-CNN+VGG M and Faster R-CNN+VGG M) in the most right plot of the Figure~\ref{fig:param_n}. This evaluation is conducted to verify our assumption that the performance of the fusion among high performers may not provide much change in performance as the shape parameter $n$ of the theoretical detector change. This additional evaluation also supports that using the ideal detector degrades detection accuracy for every object category.

\begin{figure*}[t]
    \centering
        \subfloat[bottle category] {
    	\includegraphics[trim = 120mm 60mm 85mm 50mm,clip=true,width=\textwidth]{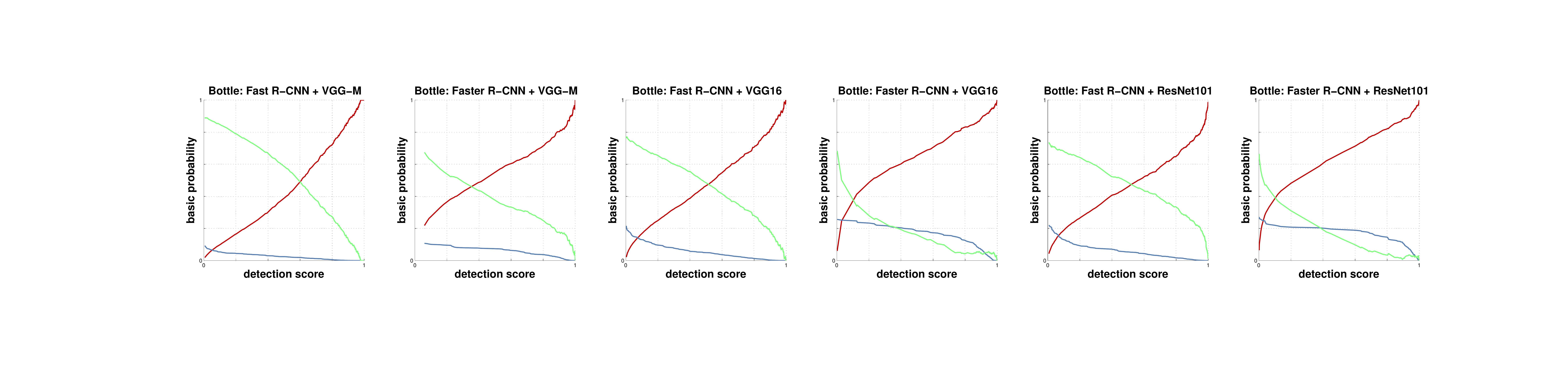}}\\
        \subfloat[car category] {
    	\includegraphics[trim = 120mm 60mm 85mm 50mm,clip=true,width=\textwidth]{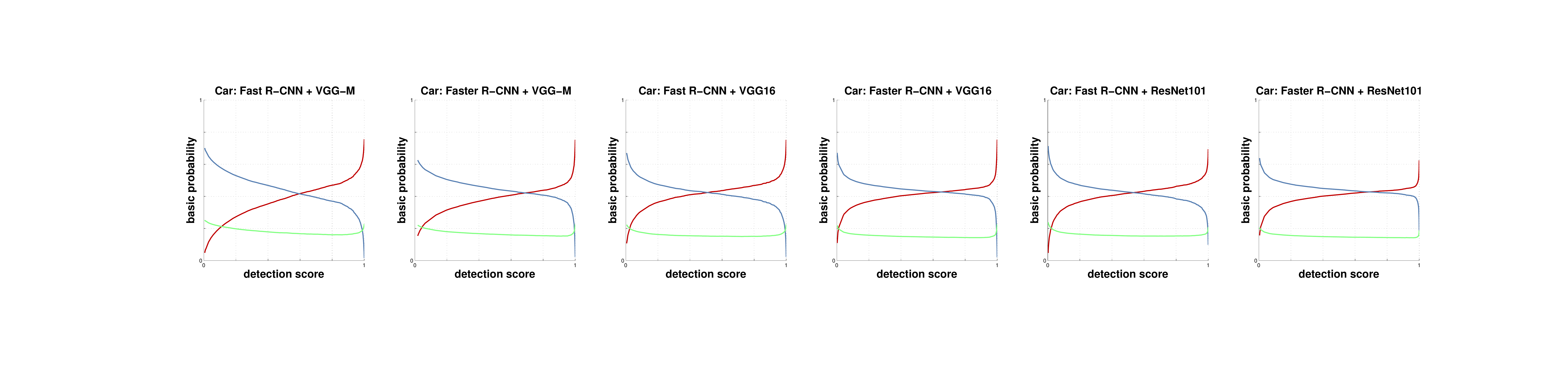}}\\
        \subfloat[person category] {
    	\includegraphics[trim = 120mm 60mm 85mm 50mm,clip=true,width=\textwidth]{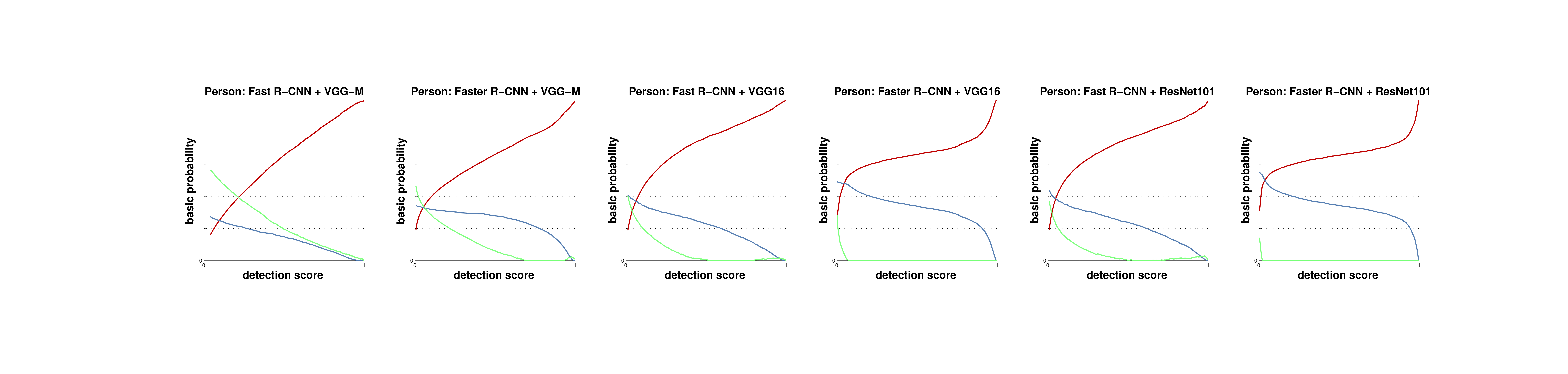}}\\
    \caption{{\bf Confidence models for several object categories on PASCAL VOC 07.} For three categories (bottle, car, and person), confidence models containing basic probability distribution varied according to a detection score are shown. \textcolor{myred}{Red}, \textcolor{myblue}{blue}, and \textcolor{mygreen}{green} curves indicate probability of \emph{target}, \emph{non-target}, and \emph{intermediate} hypotheses, respectively.}
    \label{fig:bpd_deep}
\end{figure*}

\subsection{Does Weak Detector Provide Complementary Information in the Fusion?}

\begin{table}[t]
\caption{{\bf Comparison of fusion performance} with respect to the combination of multiple detectors.}
\setlength{\tabcolsep}{5pt}
\renewcommand{\arraystretch}{1.4}
\centering
\begin{tabular}{l|ccccccc}
\specialrule{.15em}{.05em}{.05em}
\# of detectors & 2 & 3 & 4 & 5 & 6 & 7 & 8 \\\specialrule{.15em}{.05em}{.05em}
ARL & .295 & .319 & .325 & $\cdot$ & $\cdot$ & $\cdot$ & $\cdot$ \\
VOC 07 & .545 & .547 & .548 & .548 & .552 & .553 & .553\\
VOC 07 w/ R-CNNs & .742 & .750 & .752 & .759 & .760 & $\cdot$ & $\cdot$ \\\specialrule{.15em}{.05em}{.05em}
\end{tabular}
\label{tab:eval_best_comb}
\end{table}

To investigate whether (and to what degree) complementary information is provided by individual detectors in DBF, mAP is calculated while varying the number of individual detectors used in fusion. Results, shown in Table~\ref{tab:eval_best_comb} for all experiments (except PASCAL VOC 12 with R-CNNs), illustrate the performance improves as the number of detectors increases, at a decreasing rate. We observe that even the weakest detector provides complementary information to the detector pool.

\section{Further Analyses \& Discussions}
\label{sec:discuss}

\subsection{Detection Error Analysis}

We have analyzed the types of detection errors from individual detectors and DBF by using tools introduced by Hoiem et al.~\cite{DHoiemECCV12} to understand how the error types of DBF changes. Top-ranked false positives from the third evaluation setting (i.e. PASCAL VOC 07 with R-CNNs) are analyzed and categorized into four types in Figure~\ref{fig:diag_fp}. The four types of false positives (FP) are 1) poor localization (Loc, a detection with an IoU overlap with the correct class between 0.2 and 0.5, or a duplicate), 2) confusion with similar classes (Sim), 3) Confusion with dissimilar object categories (Oth), and 4) confusion with background (BG). 

We have observed that DBF reduces `Loc' errors from individual detectors. This reduction may be triggered by two factors. First, more accurate detections with higher IoU with groundtruth bounding boxes may be obtained by high performers with relatively high probabilities. Accordingly, relatively high values of precision are assigned to these detections, which are likely to be chosen as the final detections after NMS. Second, adopting bounding box refinement (introduced in section ~\ref{ssec:implementation_details}) is beneficial in computing more accurate bounding boxes.

Additionally, it is also observed that a portion of `Sim' errors is increased. It may be caused by more than two individual detectors being frequently confused with similar classes due to their similar appearances. (E.g. motorbike and bicycle.)

\subsection{Confidence Models}

In Figure~\ref{fig:bpd_deep}, the confidence models of R-CNNs for a few sample categories (bottle, car, and person) are shown, to understand variations of basic probability distribution w.r.t. varying detection scores. Generally, when a detection score is high, its corresponding probability of $T$ and $\neg T$ are relatively high and low, respectively while its tendency is reversed for low detection scores.

In terms of the probability of $I$, we have observed three characteristics. First, the variation is large for a certain category wherein detectors generally provide relatively low detection accuracy ({\it low belief for the object category}). For example, the probability of $I$ of the 'bottle' category changes more significantly than that of other categories, such as `car', and `person'. Second, for weak performers ({\it low belief for the detector}), we can observe that the lower the performance the larger the variation. For example, in the `person' category, the variations of the probability of $I$ of the left three detectors are larger than that of the other three detectors. The three detectors in the left provide worse detection accuracy than the others. Lastly, low detection scores have relatively high probabilities of $I$ ({\it low belief in a detection with low detection score}).

\subsection{PASCAL VOC Dataset Partition}

For PASCAL VOC evaluations, we use two datatset partitions. The first partition ({\tt train}/{\tt val}/{\tt test}) is used to avoid overfitting while optimizing both training detectors and building confidence models. However, for individual detectors, detection accuracy achieved by using this partition is not comparable with that reported in the original literature. This is because, in the original literature, the larger training dataset ({\tt trainval}) is used to train the detector. 

For the third setting, we use the second partition ({\tt trainval}/{\tt trainval}/{\tt test}) allowing overfitting. As in Table~\ref{tab:retraining_verification}, the accuracy achieved by retraining from the original implementation is very close to that reported in the literature. This verifies that the way of retraining detectors is correct. In Table~\ref{tab:different_partitions}, DBF provides enhanced performance over the best individual detector by .039 and .021 points in mAP for the first and second partitions, respectively. Significant performance degradation by overfitting has not been observed.

\begin{table}[t]
\caption{{\bf Comparison of detection accuracy reported in the literature and achieved by our retraining models.} {\bf S}, {\bf M}, and {\bf L} refer VGG M, VGG 16, and ResNet 101, respectively.}
\setlength{\tabcolsep}{3.0pt}
\renewcommand{\arraystretch}{1.4}
\centering
\begin{tabular}{c|c|cccccc}
\specialrule{.15em}{.05em}{.05em}
\multirow{2}{5em}{dataset} & \multirow{2}{5em}{literature/\\ours} & \multicolumn{3}{c}{Fast R-CNN} & \multicolumn{3}{c}{Faster R-CNN} \\
 & & +{\bf S} & +{\bf M} & +{\bf L} & +{\bf S} & +{\bf M} & +{\bf L} \\\specialrule{.15em}{.05em}{.05em}
\multirow{2}{5em}{VOC07} & literature & .592~\cite{RGirshickICCV15} & .669~\cite{RGirshickICCV15} & $\cdot$ & $\cdot$ & .699~\cite{SRenPAMI17} & $\cdot$ \\
& ours & .606 & .686 & .718 & .607 & .693 & .739 \\\hline
\multirow{2}{5em}{VOC12} & literature & $\cdot$ & .657~\cite{RGirshickICCV15} & $\cdot$ & $\cdot$ & .670~\cite{SRenPAMI17} & $\cdot$ \\
& ours & .580 & .666 & .687 & .568 & .673 & .727 \\\specialrule{.15em}{.05em}{.05em}
\end{tabular}
\label{tab:retraining_verification}
\end{table}

\begin{table}[t]
\caption{{\bf Different dataset partitions.} In Table, partition indicates datasets used for detector training/confidence model building.}
\setlength{\tabcolsep}{4.5pt}
\renewcommand{\arraystretch}{1.4}
\centering
\begin{tabular}{c|c|cccccc}
\specialrule{.15em}{.05em}{.05em}
\multirow{2}{5em}{partition} & \multirow{2}{2em}{DBF} & \multicolumn{3}{c}{Fast R-CNN} & \multicolumn{3}{c}{Faster R-CNN} \\
 & & +{\bf S} & +{\bf M} & +{\bf L} & +{\bf S} & +{\bf M} & +{\bf L} \\\specialrule{.15em}{.05em}{.05em}
{\tt train}/{\tt val} & .709 & .546 & .631 & .644 & .551 & .621 & .670 \\
{\tt trainval}/{\tt trainval} & .760 & .606 & .686 & .718 & .607 & .693 & .739 \\\specialrule{.15em}{.05em}{.05em}
\end{tabular}
\label{tab:different_partitions}
\end{table}

\section{Conclusions}
\label{sec:concl}

A novel fusion method, referred to as Dynamic Belief Fusion (DBF), is proposed to improve upon current late fusion methods in the task of object detection.

For object detection, we consider three hypotheses \emph{target}, \emph{non-target}, and \emph{intermediate state}, where the last one indicates ambiguity between \emph{target} and \emph{non-target}. DBF assigns basic probabilities to the three hypotheses estimated from current detection score and the confidence models built on the previously computed precision-recall curves on a validation image set. In order to properly assign a probability to \emph{intermediate states}, the PR curve of a theoretical detector is artificially modeled and the difference on the PR curves between the individual detector and the theoretical detector at a given recall value is used to represent a degree of belief for the intermediate hypothesis. Dempster's combination rule is used to combine the basic probabilities of detection results from different detectors.

The extensive experimental results demonstrate that DBF is superior to all baseline fusion approaches and all individual detectors with respect to mean average precision (mAP). The enhanced performance of DBF over DST-based fusion methods incorporating a fixed level of probability into the fusion process clearly shows the robustness of the dynamic basic probability assignment. Also, the superior performance of DBF to Bayesian fusion strongly supports the use of an intermediate belief state, which was achieved in this context via the instantiation of a theoretical detector in conjunction with individual detectors. Finally, DBF provides enhanced fusion performance over the best detector as well as all the individual detectors in the fusion pool.


\ifCLASSOPTIONcaptionsoff
  \newpage
\fi



\bibliographystyle{IEEEtran}
\bibliography{egbib.bib}

\end{document}